\documentclass[lettersize,journal]{IEEEtran}
\usepackage{amsmath,amsfonts}
\usepackage{amsthm}
\usepackage{algorithmic}
\usepackage[ruled,linesnumbered]{algorithm2e}
\usepackage{array}
\usepackage[caption=false,font=footnotesize,labelfont=rm,textfont=rm]{subfig}
\usepackage{textcomp}
\usepackage{stfloats}
\usepackage{url}
\usepackage{verbatim}
\usepackage{graphicx}
\usepackage{cite}
\usepackage{color}
\usepackage{multirow}
\usepackage{booktabs}
\usepackage[pagebackref=false,breaklinks=false,linkcolor=red,anchorcolor=black, citecolor=green,colorlinks,bookmarks=true]{hyperref}
\begin{document}

\title{Discovering and Explaining the Non-Causality of Deep Learning in SAR ATR}
\author{Weijie Li, Wei Yang, Li Liu,~\IEEEmembership{Senior member,~IEEE}, Wenpeng Zhang, Yongxiang Liu,~\IEEEmembership{member,~IEEE}
\thanks{This work was supported by the National Key Research and Development Program of China (No. 2021YFB3100800), the National Natural Science Foundation of China (No. 61871384, 61901487, 61901498, and 61921001), and the Science and Technology Innovation Program of Hunan Province (No. 2022RC1092). \emph{(Corresponding author: Wei Yang.)}}
\thanks{The authors are with the College of Electronic Science and Technology, National University of Defense Technology, Changsha 410073, China (e-mail: lwj2150508321@sina.com, yw850716@sina.com, dreamliu2010@gmail.com, zhangwenpeng08@nudt.edu.cn, and lyx\_bible@sina.com).
}}

\markboth{In preparation for submission}%
{Li \MakeLowercase{\textit{et al.}}: A Sample Article Using IEEEtran.cls for IEEE Journals}


\maketitle

\begin{abstract}
\textcolor{blue}{This is the pre-acceptance version, to read the final version please go to IEEE Geoscience and Remote Sensing Letters on IEEE Xplore.}
In recent years, deep learning has been widely used in SAR ATR and achieved excellent performance on the MSTAR dataset. However, due to constrained imaging conditions, MSTAR has data biases such as background correlation, \emph{i.e.}, background clutter properties have a spurious correlation with target classes. Deep learning can overfit clutter to reduce training errors. Therefore, the degree of overfitting for clutter reflects the non-causality of deep learning in SAR ATR. Existing methods only qualitatively analyze this phenomenon.
In this paper, we quantify the contributions of different regions to target recognition based on the Shapley value. The Shapley value of clutter measures the degree of overfitting. Moreover, we explain how data bias and model bias contribute to non-causality. Concisely, data bias leads to comparable signal-to-clutter ratios and clutter textures in training and test sets. And various model structures have different degrees of overfitting for these biases. The experimental results of various models under standard operating conditions on the MSTAR dataset support our conclusions.
\end{abstract}


\section{Introduction}
\IEEEPARstart{S}{YNTHETIC} Aperture Radar (SAR) Automatic Target Recognition (ATR) is an essential branch of SAR image interpretation with significant promise for military and civilian applications\cite{blasch2020review, kechagias2021automatic}. As deep learning has developed rapidly in this field, deep learning-based methods have achieved a recognition rate of over 99\% percent for the benchmark dataset under Standard Operating Conditions (SOC)\cite{blasch2020review, kechagias2021automatic}.

\begin{figure*}[!tb]
\centering
\includegraphics[width=18cm]{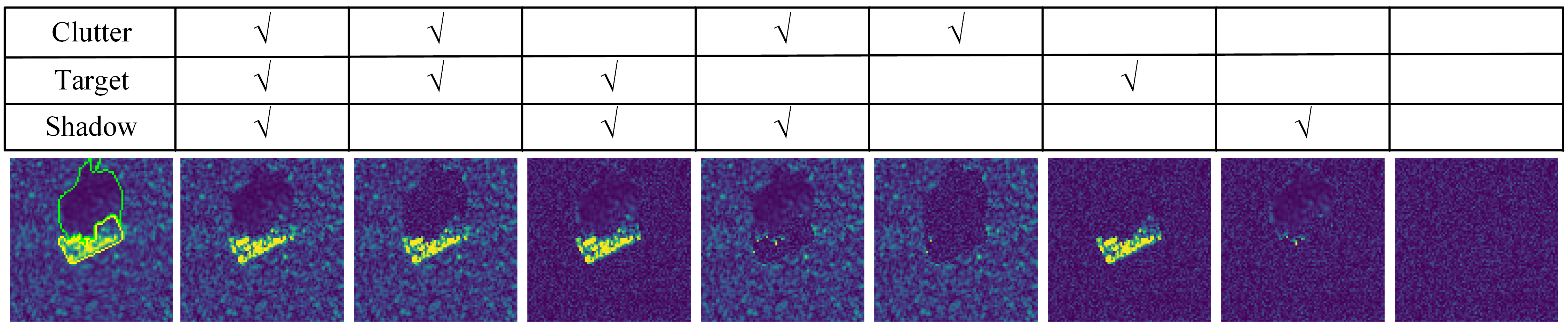}
\caption{Ablation of SAR image regions (check mark represents the reserved areas). The target is positioned in the center and has a strong magnitude. The shadow is a dark region behind the target due to the occlusion in the radar line of sight. Clutter is a disorganized region around the target with different textures.}
\label{fig_segmation}
\end{figure*}

However, impressive performance does not mean deep learning correctly solves this recognition problem. The data and model affect the deep learning-based feature representation of SAR images. At the data level, the Moving and Stationary Target Acquisition and Recognition (MSTAR) benchmark dataset has data bias\footnote{Data bias refers to spurious correlations between target classes and non-causal features due to non-random collection conditions\cite{ref61}. For example, the background contains no information about the target classes. Nevertheless, during data collection, the backgrounds of targets have differences across classes, which establishes a spurious correlation between backgrounds and target classes.} (\emph{i.e.}, data selection bias). One of the obvious biases in MSTAR is background correlation\cite{kechagias2021automatic}. This spurious correlation allows deep learning to exceed the recognition rate of random guesses when using background clutter slices without targets\cite{kechagias2021automatic, ref6, ref18}. It is apparent from this phenomenon that clutter properties are class-related (\emph{i.e.}, clutter properties correlate with target classes) between the training set and the test set. However, to the best of our knowledge, researchers have yet to analyze the impact of background correlation on deep learning quantitatively. 
At the model level, deep learning can learn feature representation in a way that is unexpected to humans\cite{ref59}. An example of model bias is texture\footnote{Texture refers to the spatial organization of a set of basic elements\cite{liu2019bow}. As shown in Fig. \ref{fig_segmation}, the scattering point amplitude has different spatial distributions in the target, clutter, and shadow regions. These spatial distributions create texture signatures of different regions in SAR images.} bias\cite{shi2020informative}, \emph{i.e.}, deep learning relies on local texture rather than shape or contour for recognition. This problem is also present in SAR ATR due to imaging characteristics. SAR imaging reflects back-scattering intensity, which lacks the complete geometric structure and color of targets. As shown in Fig. \ref{fig_segmation}, the main visual information is the scattering point distributions in the target and background regions, which reflect scattering point intensity and texture information. Therefore, deep learning can exploit the intensity and texture of background clutter.

\IEEEpubidadjcol

Previous deep learning interpretability studies of SAR ATR \cite{ref29,ref25,belloni2020explainability} calculated the contribution of each single-pixel point in target, clutter, and shadow regions. Moreover, the ablation studies\cite{belloni2020explainability, ref6,ref18} described the influence of regions on the recognition rate. In particular, Belloni \emph{et al.}\cite{belloni2020explainability} discussed the influence of different region permutations on the recognition rate of each class by ablation studies on target, clutter, and shadow regions. The above studies show that deep learning relies on target signatures and background clutter. The overfitting for background clutter reflects the incorrect feature representation and non-causality of deep learning. However, existing studies\cite{belloni2020explainability,ref29,ref25} focus on pixel-level contribution values and cannot quantitatively reflect the overall impact of different regions. It is also difficult to conclude intuitively from the recognition rates \cite{belloni2020explainability,ref18} of target, clutter, and shadow permutations. Therefore, we improve how the saliency map divides images and calculates contributions. We analyze the contributions of different regions based on the Shapley value to provide quantitative indicators and clear conclusions.

This paper first analyzes the contributions and interactions of targets, clutter, and shadow regions during training. The contribution of clutter can be used as a quantitative indicator of the non-causality of deep learning. Next, we explain how data and model biases affect causality. The comparable signal-to-clutter ratios (SCR) and clutter texture establish a spurious correlation between target classes and background clutter. Deep learning can overfit these differences to reduce training errors. A re-weighting method is proposed to verify whether SCR is the cause of overfitting. Finally, we analyze the texture bias of the model structure with different parameters. 
The common phenomenon of various models in the MSTAR dataset supports our analysis and conclusions. The main contributions of this paper are as follows. Our code is available at \url{https://github.com/waterdisappear/Data-Bias-in-MSTAR}.

\begin{enumerate}
\item{We use the Shapley value and bivariate Shapley interactions to reflect the contribution and interactions of regions. These quantitative metrics can be used to analyze the feature representation of deep learning during training.}
\item{Quantitative metrics allow us to analyze factors that impact deep learning causality. By intervening in the clutter SCR and model parameters, the quantitative metrics can clearly show the factors that influence overfitting for clutter.}
\end{enumerate} 

\section{Methodology}
\subsection{Shapley Value}
Shapley value\cite{shapley} is a method for calculating the player contribution solution in a cooperative game. This unique solution satisfies the four properties and is considered a fair attribution method\cite{zhang2021interpreting, sundararajan2020many}. Consider $N$ is a set of all players with $2^{\left| N \right|}$ possible subsets. A game is a function $v:2^{\left| N \right|} \rightarrow \mathbb{R}$ that maps a subset to a real number. The Shapley value of $i$-th player is given by:
\begin{equation}\label{eq_Shapley}
\phi(i|N)=\sum_{S\subseteq N\backslash \{i\}} \frac{\left| S \right|! (\left| N \right|-\left| S \right|-1)!}{\left| N \right|!}\left[v(S\cup\{i\})-v(S)\right],
\end{equation}
where $S$ is a subset of $N\backslash \{i\}$, $\left| \cdot \right|$ is the number of elements in the set. The Shapley value $\phi(i|N)$ indicates whether $i$-th player has a positive or negative impact on the outcome.

There are many variants for computing Shapley values in deep learning\cite{lundberg2017unified,sundararajan2020many}, and we use the Baseline Shapley (BShap)\cite{sundararajan2020many}. The BShap is also the unique solution method that satisfies certain properties\cite{sundararajan2020many}:
\begin{equation}\label{eq_BShap}
v(S)=f(x_S;\tilde{x}_{N\backslash S}),
\end{equation}
where $x_S;\tilde{x}_{N\backslash S}$ means that $S$ part of input $x$ retains its original value, while other parts are replaced with baseline $\tilde{x}$. We set $f(\cdot)$ as the model classification score corresponding to the true class before Softmax\footnote{Since deep learning has saturated performance under SOC, we are more interested in the contribution of different regions to correct recognition during training. In addition, analyzing the causes of incorrect recognition in tests is also essential.}.

\subsection{Bivariate Shapley Interaction}
The Bivariate Shapley Interaction (BSI)\cite{zhang2021interpreting} is a way to measure the additional benefit of a coalition $S_{ij} = \{i,j\}$ of two players $i$ and $j$. The contribution of a coalition is often different from the sum of the individual players due to the influence of player interactions. For example, the shadow region responds to target shape information under specific sensor conditions and can form a coalition with the target region\cite{belloni2020explainability, choi2022fusion}. The BSI can be defined below:
\begin{equation}\label{eq_BShap}
\begin{split}
B(S_{ij})&=\phi(S_{ij}|N')-\left[\phi(i|N_i)+\phi(j|N_j)\right]\\
&=\sum_{S\subseteq N\backslash \{ i,j \} } \frac{\left| S \right|! (\left| N \right|-\left| S \right|-2)!}{(\left| N \right|-1)!}\left[\Delta f(S,i,j)\right],
\end{split}
\end{equation}
where $N'=N\backslash \{i,j\}\cup S_{ij}$, $N_i=N\backslash \{j\}$, $N_j=N\backslash \{i\}$, $\Delta f(S,i,j)=v(S\cup \{i,j\})-v(S\cup \{i\})-v(S\cup \{j\})+v(S)$. The BSI describes whether an alliance between two players benefits the outcome.

\subsection{Random SCR Re-Weighting}
The comparable SCR\footnote{Please refer to Fig. \ref{fig_class}, which provides the comparable SCR curves} across classes between the training and test sets is a brief representation of the background correlation. We propose a random SCR re-weighting method inspired by the re-weighting methods in causal inference\cite{yao2021survey}. We intend to create a similar SCR between classes to remove this spurious correlation. Our experiments show that this intervention reduces the overfitting for clutter, which shows that SCR is indeed one of the factors contributing to the background correlation. Alternatively, SCR re-weighting preserves the clutter texture, so we can further analyze the texture bias in various models.

First, the SCR of SAR images is defined as the ratio of the mean pixel value of target region $\overline{m}^{\rm{tar}}$ to the mean pixel value of clutter region $\overline{m}^{\rm{clu}}$, \emph{i.e.}, $\rm{SCR}=20\ lg(\frac{\overline{m}^{\rm{tar}}}{\overline{m}^{\rm{clu}}})$. For the input magnitude image matrix $\mathbf{X}$, $\rm{SCR}^{\prime}$ is sampled from a specific distribution\footnote{A uniform distribution is used to simulate random sampling, which keeps the SCR of different samples in the same 3 dB range. We aim to verify whether the comparable SCR leads to overfitting for clutter.}, such as a uniform distribution $U(11, 14)$. We calculate the re-weighting factor $\alpha=10^{\frac{(\rm{SCR} -\rm{SCR}^{\prime})}{20}}$ and the re-weighting image $\mathbf{X}^{\prime}$ is 
\begin{equation}\label{eq_BShap} \mathbf{X}^{\prime}=\mathbf{X} \circ (\mathbb{I}-\mathbf{K}^{\rm{clu}})+\alpha \cdot \mathbf{X} \circ \mathbf{K}^{\rm{clu}},
\end{equation}
where $\mathbf{K}^{\rm{clu}}$ is the mask matrix of the clutter region, $\mathbb{I}$ is a matrix whose elements are all one, $\circ$ is the Hadamard product.

\section{Experiments}

\begin{figure*}[!tb]
\centering
\subfloat[Shapley value.]{\includegraphics[width=5.5cm]{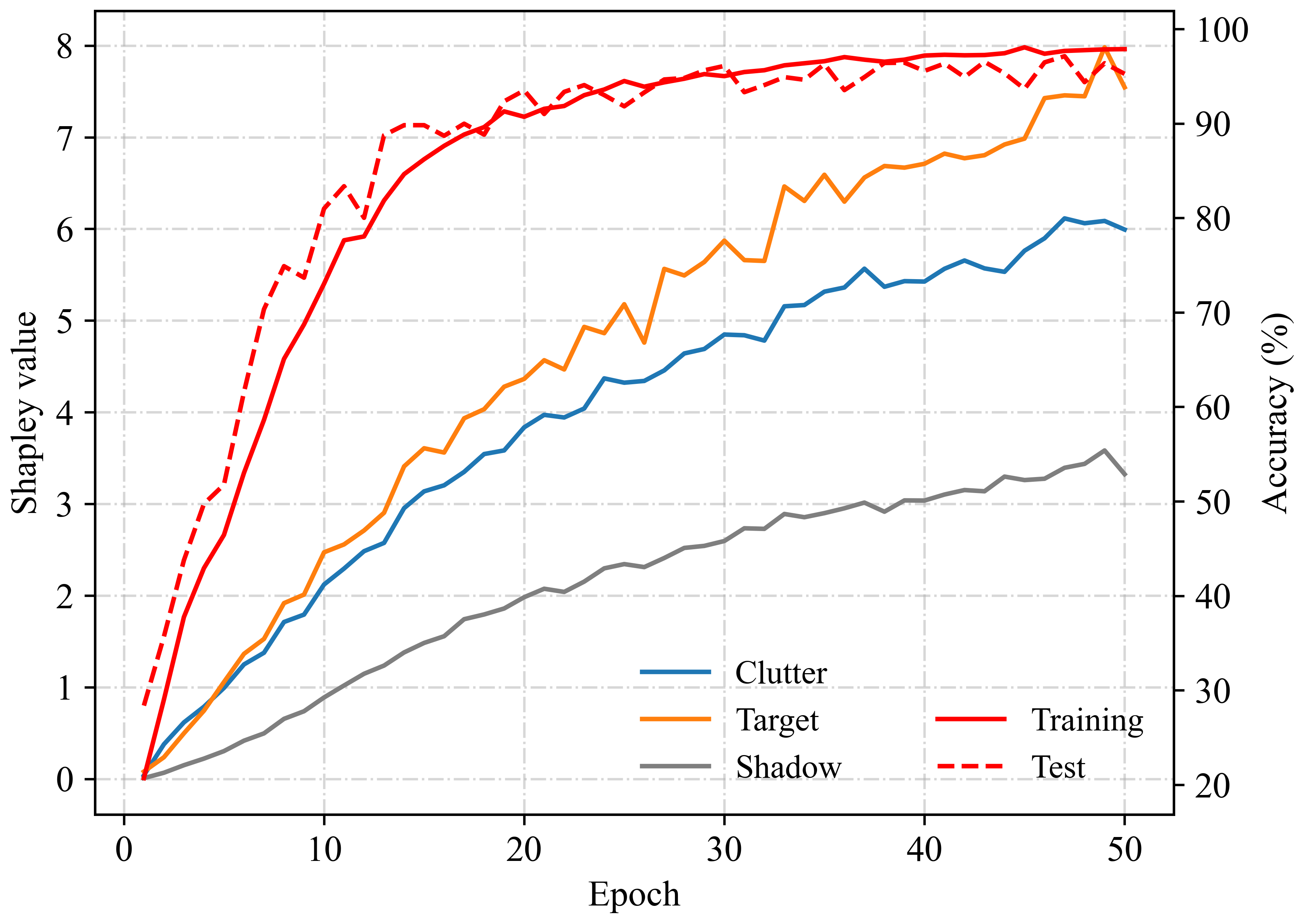}\label{model_1_shapley}}
\subfloat[Shapley value ratio.]{\includegraphics[width=5.5cm]{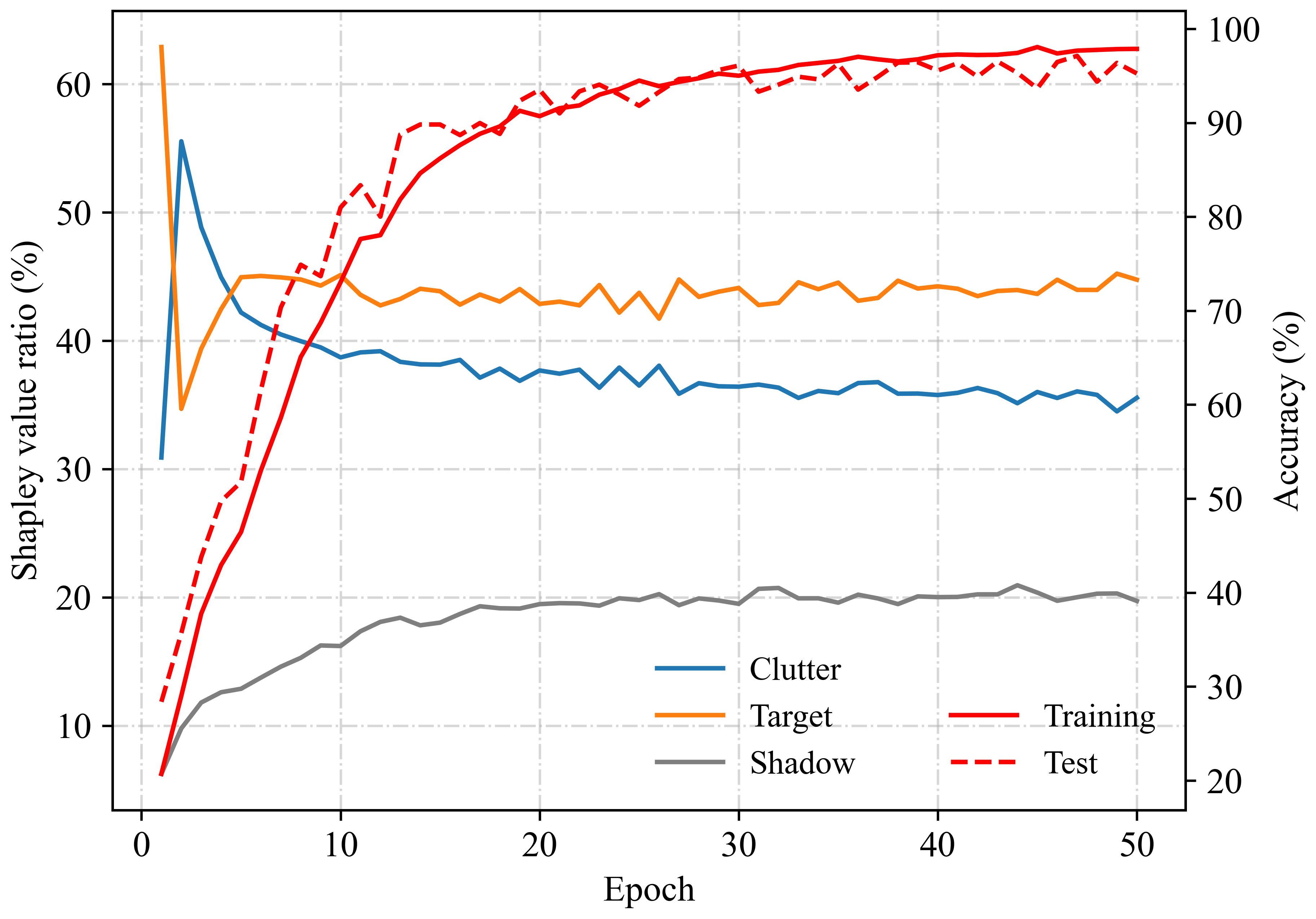}\label{model_1_shapley_ratio}}
\subfloat[Bivariate Shapley interaction.]{\includegraphics[width=5.5cm]{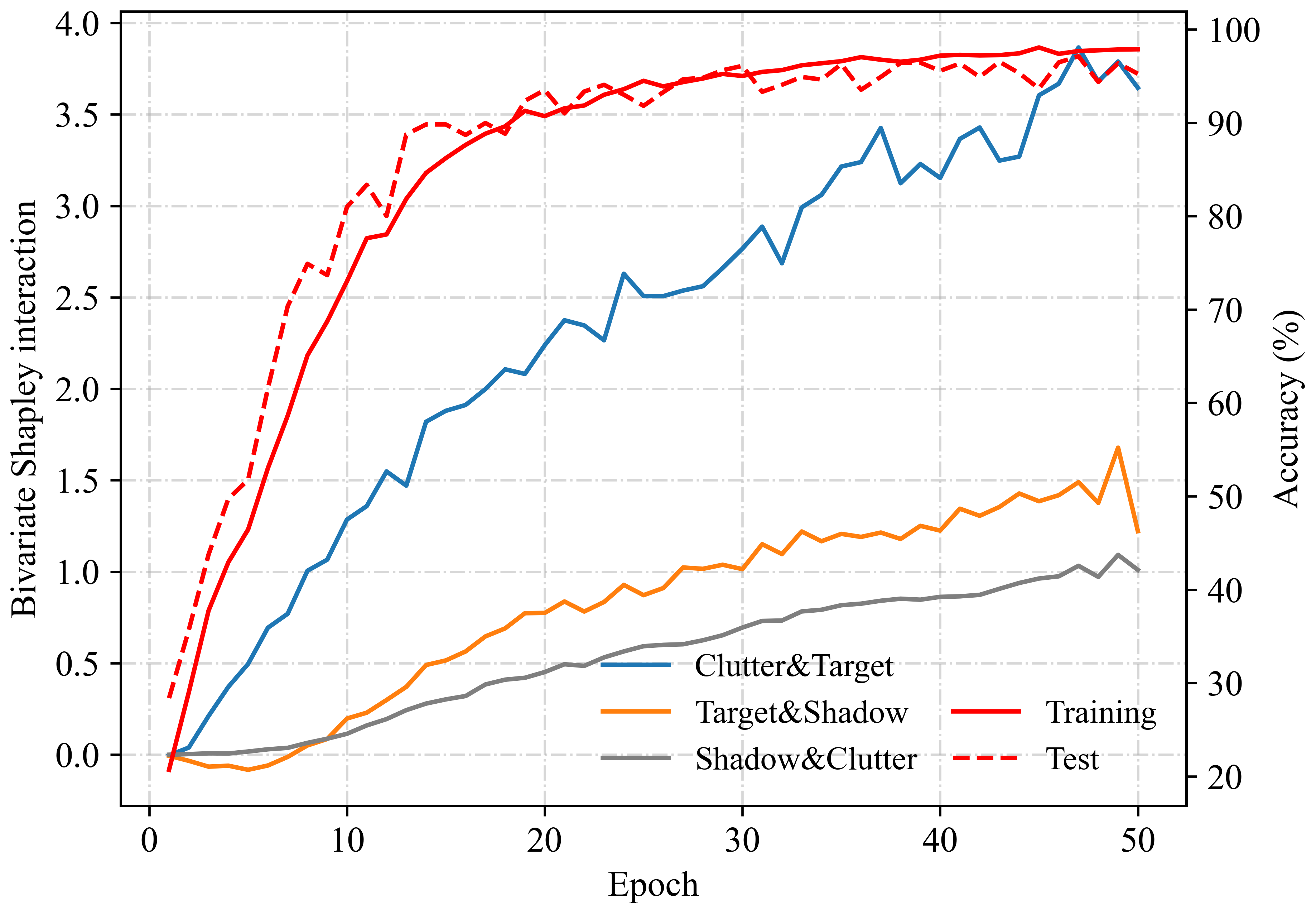}\label{model_1_bshapley}}
\caption{The training of A-ConvNet. Both Shapley value and BSI reflect that the correlations captured by deep learning contain non-causality. (Other results are available in the supporting material. The contributions of different regions show varying trends across models, but the same thing remains: clutter always occupies a certain percentage.)}
\label{model_1_sha}
\end{figure*}

\subsection{Data and Model}
MSTAR, the most cited SAR ATR dataset in recent years, was collected by Sandia National Laboratory in the 1990s\cite{blasch2020review, kechagias2021automatic}. The SAR images in JPEG format are generated by the official conversion tool\footnote{The official tool generates JPEG images by linear mapping and contrast enhancement. Since linear mapping normalizes the amplitude to [0, 255], we use SCR (a ratio) instead of amplitude to represent clutter properties. The contrast enhancement enhances the amplitude of targets and some clutter points. This method reduces A-ConvNet overfitting for clutter.} and use SARbake\footnote{We use SARbake as a benchmark since it is an open-source dataset for segmentation. As SARbake uses 3D CAD models and projection relations to determine regions, some segmentation results do not completely match the actual SAR image.} segmentation\cite{malmgren2015convolutional} as the ground truth for target, shadow, and clutter regions. SOC is the recognition of ten classes of targets\cite{kechagias2021automatic}. The training and test sets are images of different azimuth angles in a similar scene\footnote{Although the scenes are all flat grasslands, factors such as the height, sparseness, and water content of vegetation can affect the scattering properties of background clutter.}, while the depression angle differs by 2°. The benchmark models are three deep learning models (A-ConvNet\cite{chen2016target}, AM-CNN\cite{zhang2020convolutional}, MVGGNet\footnote{MVGGNet used optical pre-training weights in the original paper. Considering our discussion of parameters, the optical pre-training weights are used for models with parameters larger than MVGGNet. Previous work\cite{ref29} has shown that convolution kernels with pre-training have better texture and shape properties than direct training in the MSTAR dataset. We find this to be critical for ConvNeXtTiny. Optical pre-training weights reduced its overfitting for clutter and improved the test set accuracy from 55\% to 98\%.}\cite{zhang2020fec}) in SAR ATR and other models (EfficientNet\cite{tan2019efficientnet}, ResNet\cite{he2016deep} and ConvNeXt\cite{liu2022convnet}) in computer vision.

\subsection{Implementation Detail}
The total score of BShap is $f(x)-f(\emptyset)=\sum_{i} \phi(i|N)$. Therefore the baseline setting of Shapley values (\emph{i.e.}, absence of input variables) is essential. Belloni \emph{et al.}\cite{belloni2020explainability} set the absence value to zero to calculate the recognition rate change. Considering the dark shadow region\cite{belloni2020explainability, ref29}, the absolute values of a Gaussian distribution $N(0, 0.1)$ are used in Fig. \ref{fig_segmation}. This distribution differs from the target, clutter, and shadow areas to ensure that the total score reflects the differences.
In the experiments, zero or noise baselines do not guarantee that $f(\emptyset)$ is zero across classes. Moreover, various models have different classification scores $f(x)$. These cause the total score $f(x)-f(\emptyset)$ to vary across models, classes, and baselines. Therefore, we use the Shapley value ratio\footnote{Due to the negative Shapley values, the Shapley value ratio is calculated by $\frac{\phi(i|N)}{\sum_{i} abs(\phi(i|N))}$, where $abs(\cdot)$ is the absolute value.} to reflect the relative values. The Shapley values for the following experiments are the average of five replicate experiments, and Fig. \ref{model_1_sha} is the result of one.

\subsection{Discovering the Non-Causality}

\begin{table*}[!tb]
\centering
\caption{Shapley Value Ratio and Bivariate Shapley Interaction of Models}
\label{table0}
\renewcommand\arraystretch{1.3}
\resizebox{\linewidth}{!}{%
\begin{tabular}{cccccccccc} \toprule
\multirow{2}{*}{Model} & \multicolumn{3}{c}{Region (SVR)} & \multicolumn{3}{c}{Coalition (BSI)} & \multicolumn{2}{c}{Accuracy } & \multirow{2}{*}{Params} \\ \cmidrule(lr){2-4}\cmidrule(lr){5-7}\cmidrule(lr){8-9}
 & Clutter & Target & Shadow & Clutter\&Target & Target\&Shadow & Shadow\&Clutter & Training & Test & \\ \midrule
A-ConvNet\cite{chen2016target} & 35.91\% (5.80) & \textbf{44.52\% (7.19)} & 19.57\% (3.16) & \textbf{4.48} & 1.57 & 1.10 & 97.81\% & 95.70\% & 0.30M \\
AM-CNN\cite{zhang2020convolutional} & 27.08\% (2.71) & \textbf{53.23\% (5.32)} & 19.69\% (1.97) & \textbf{3.86} & 2.87 & 1.20 & 100.0\% & 97.17\% & 2.60M \\
EfficientNet-B0\cite{tan2019efficientnet} & 31.03\% (4.49) & \textbf{40.79\% (5.90)} & 28.18\% (4.08) & \textbf{3.05} & 1.75 & 1.93 & 99.40\% & 92.83\% & 4.02M \\
EfficientNet-B1\cite{tan2019efficientnet} & 33.29\% (4.63) & \textbf{38.16\% (5.31)} & 28.55\% (3.97) & \textbf{3.29} & 1.95 & 1.50 & 99.51\% & 93.06\% & 6.53M \\
MVGGNet\cite{zhang2020fec} & \textbf{\textbf{34.21\% (8.09)}} & 33.02\% (7.81) & 32.77\% (7.75) & 4.16 & 5.11 & \textbf{\textbf{5.15}} & 95.98\% & 95.83\% & 16.81M \\
ResNet34\cite{he2016deep} & \textbf{39.39\% (4.94)} & 34.53\% (4.33) & 26.08\% (3.27) & \textbf{3.06} & 2.29 & 2.10 & 99.56\% & 95.81\% & 21.29M \\
ResNet50\cite{he2016deep} & \textbf{37.80\% (4.73)} & 36.46\% (4.57) & 25.74\% (3.22) & \textbf{2.78} & 1.46 & 1.94 & 99.85\% & 98.00\% & 23.53M \\
ConvNeXtTiny\cite{liu2022convnet} & \textbf{49.69\% (5.38)} & 27.35\% (2.96) & 22.96\% (2.49) & \textbf{2.11} & 0.40 & 1.34 & 99.84\% & 98.19\% & 27.83M \\ \bottomrule
\end{tabular}
}
\end{table*}

\begin{figure*}[!tb]
\centering
\subfloat[A-ConvNet]{\includegraphics[width=4.5cm]{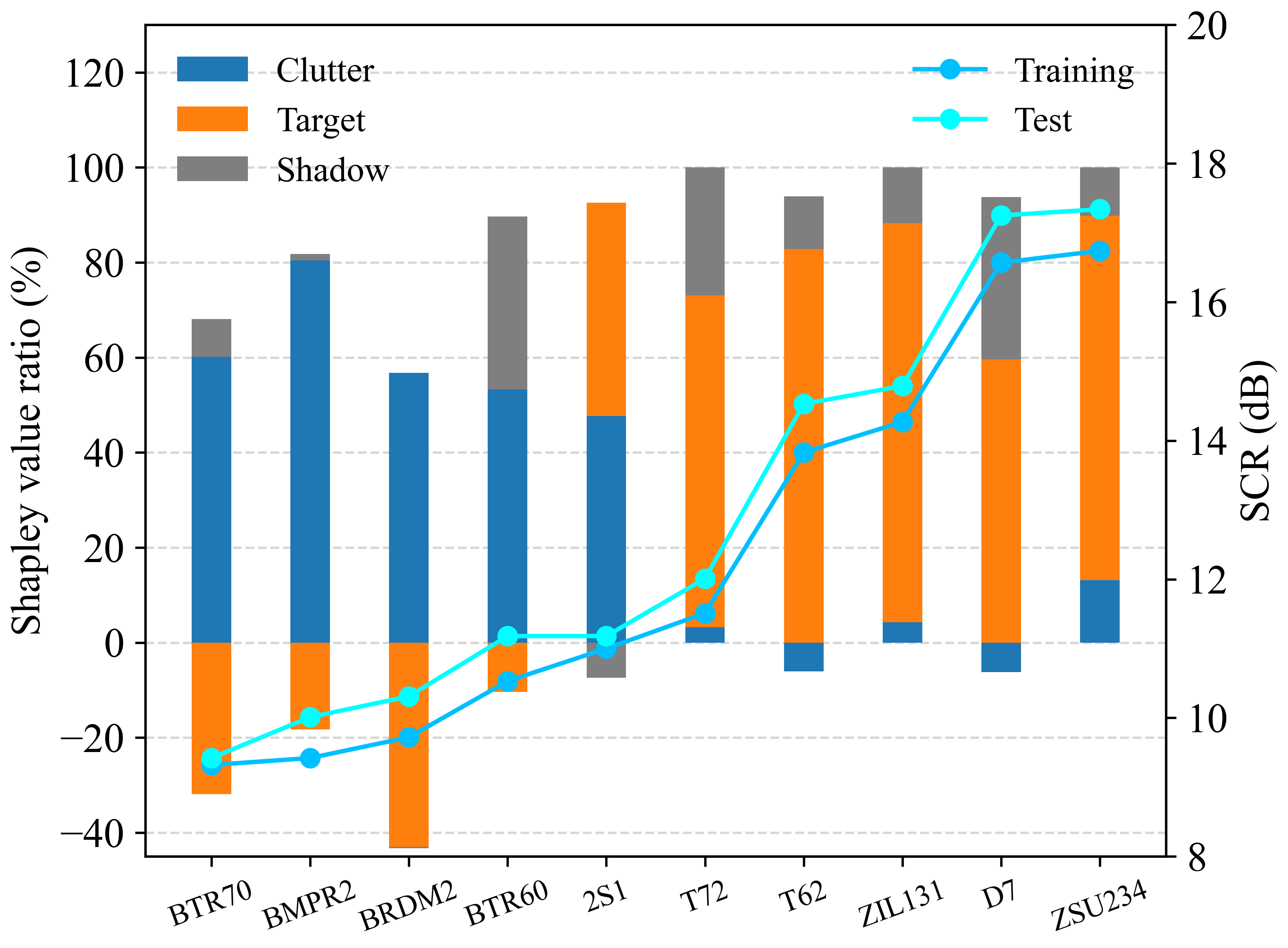}}
\subfloat[AM-CNN]{\includegraphics[width=4.5cm]{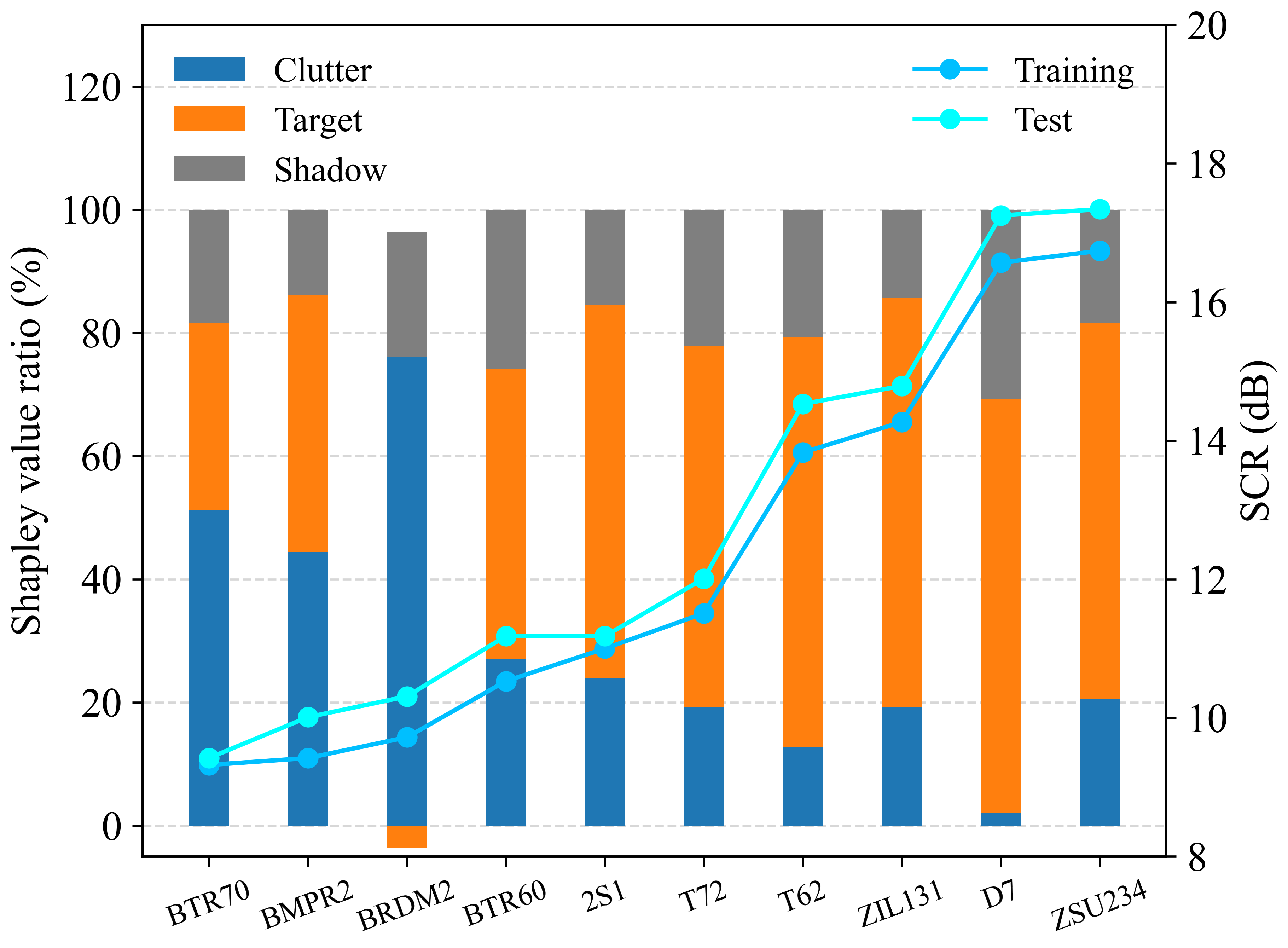}}
\subfloat[EfficientNet-B0]{\includegraphics[width=4.5cm]{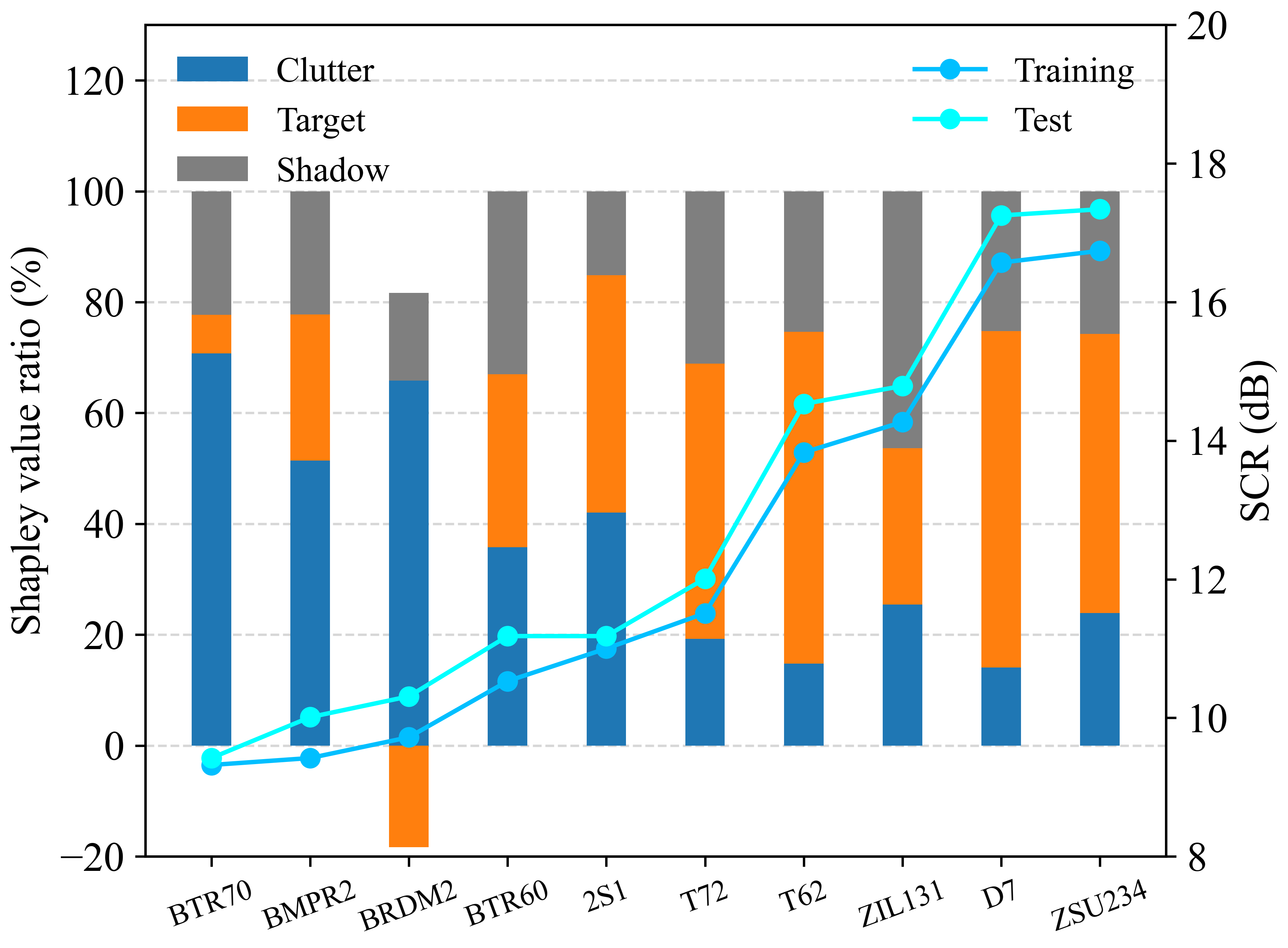}}
\subfloat[EfficientNet-B1]{\includegraphics[width=4.5cm]{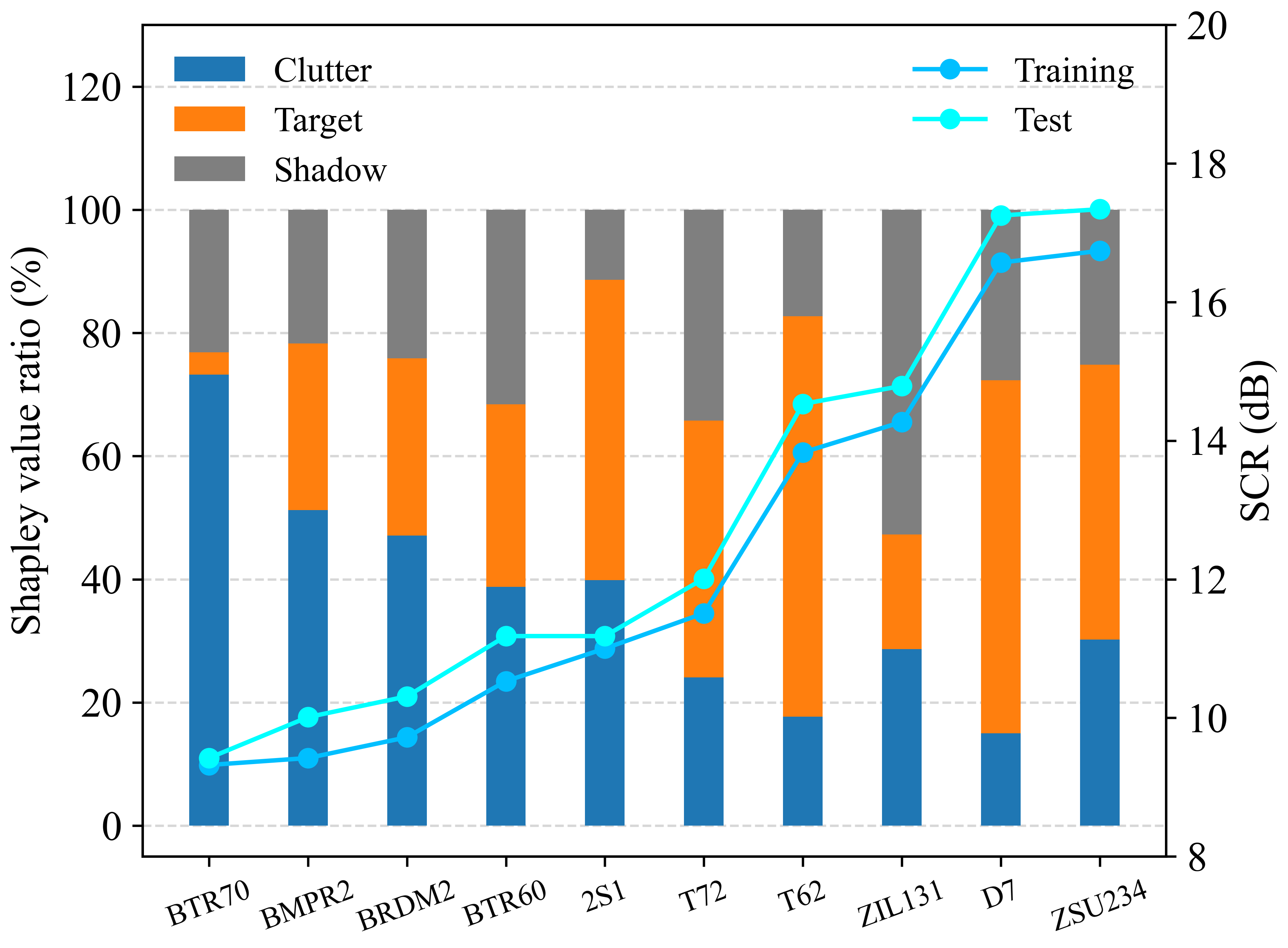}}
\vfil
\subfloat[MVGGNet]{\includegraphics[width=4.5cm]{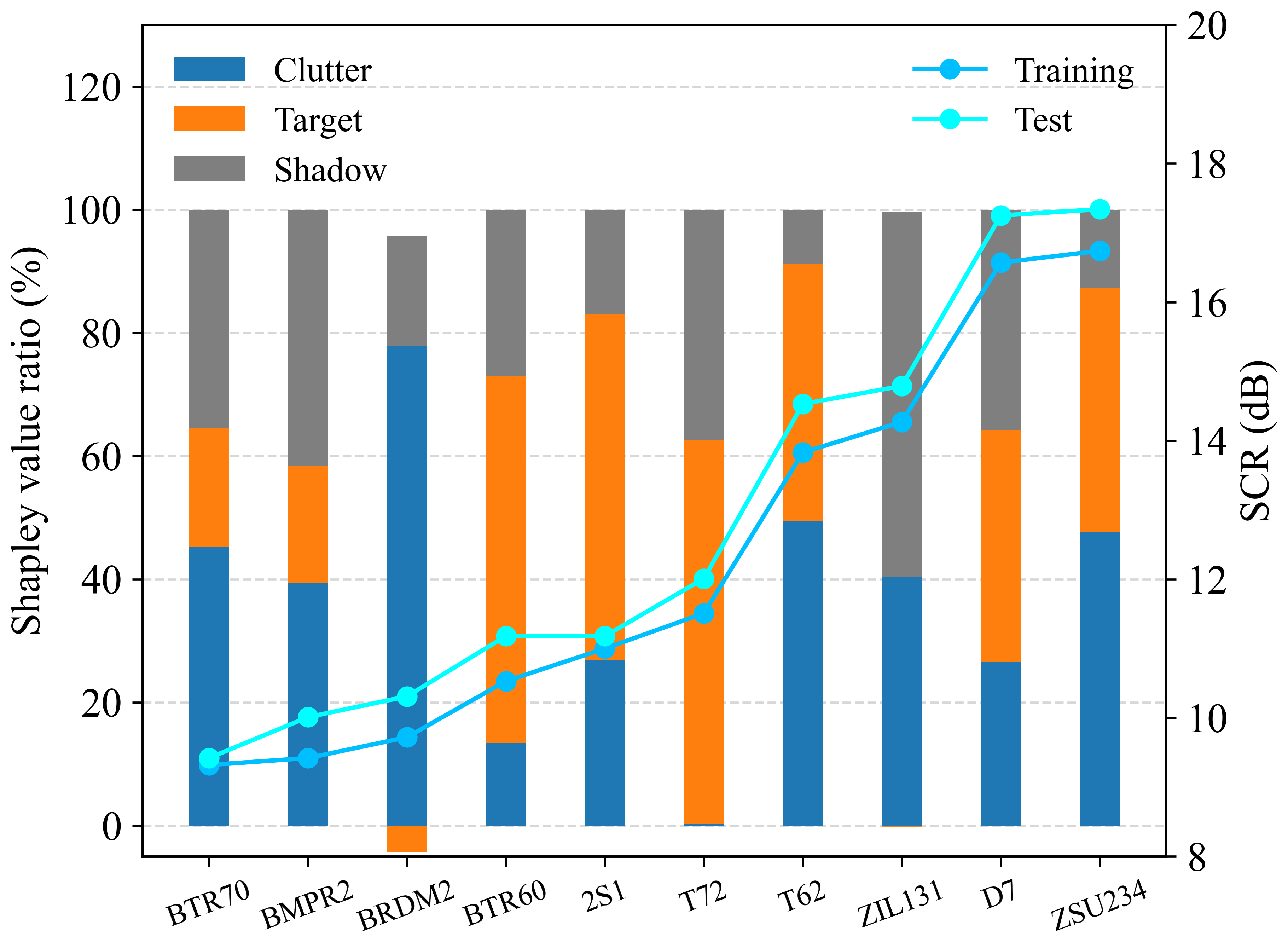}}
\subfloat[ResNet34]{\includegraphics[width=4.5cm]{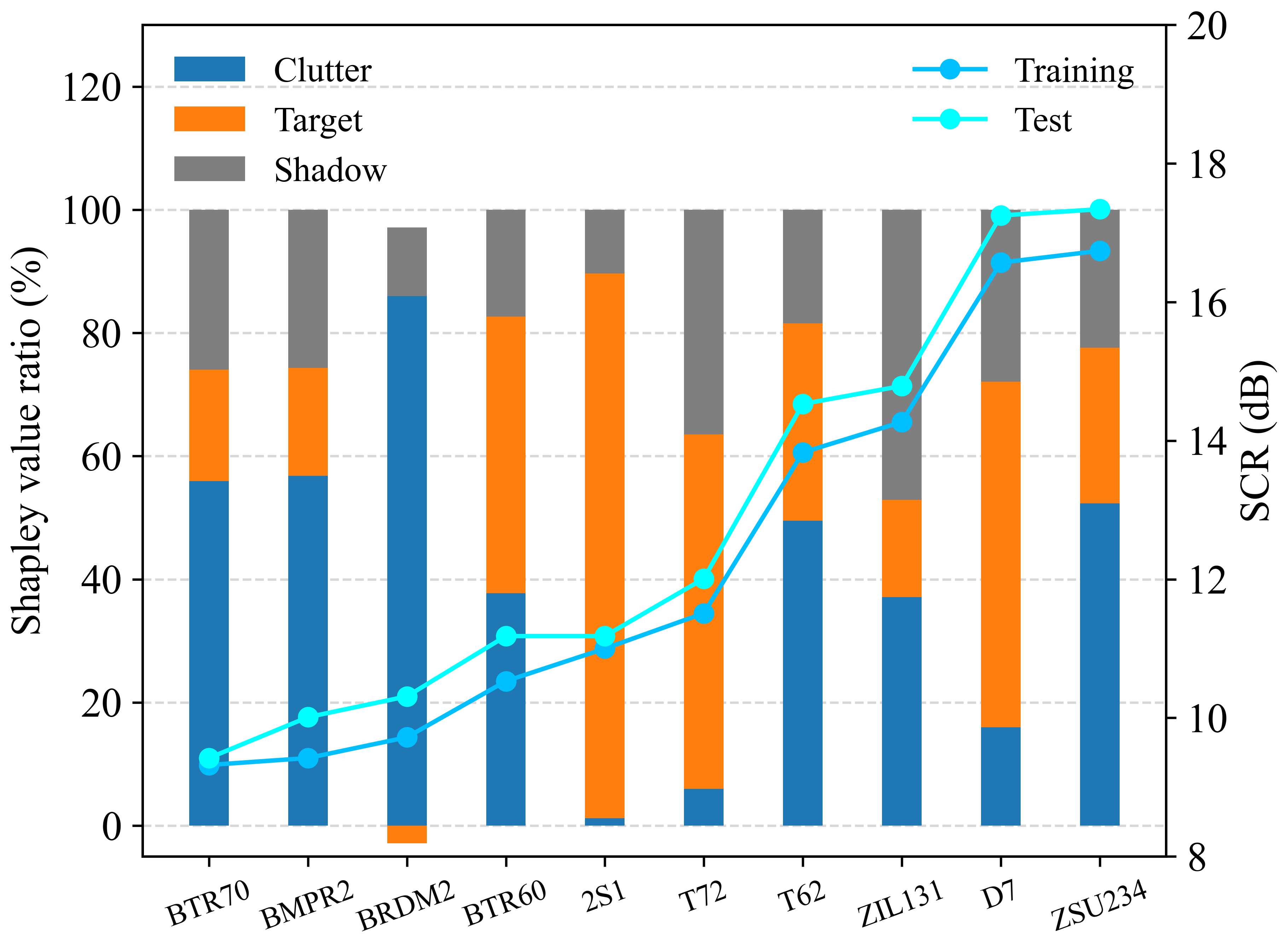}}
\subfloat[ResNet50]{\includegraphics[width=4.5cm]{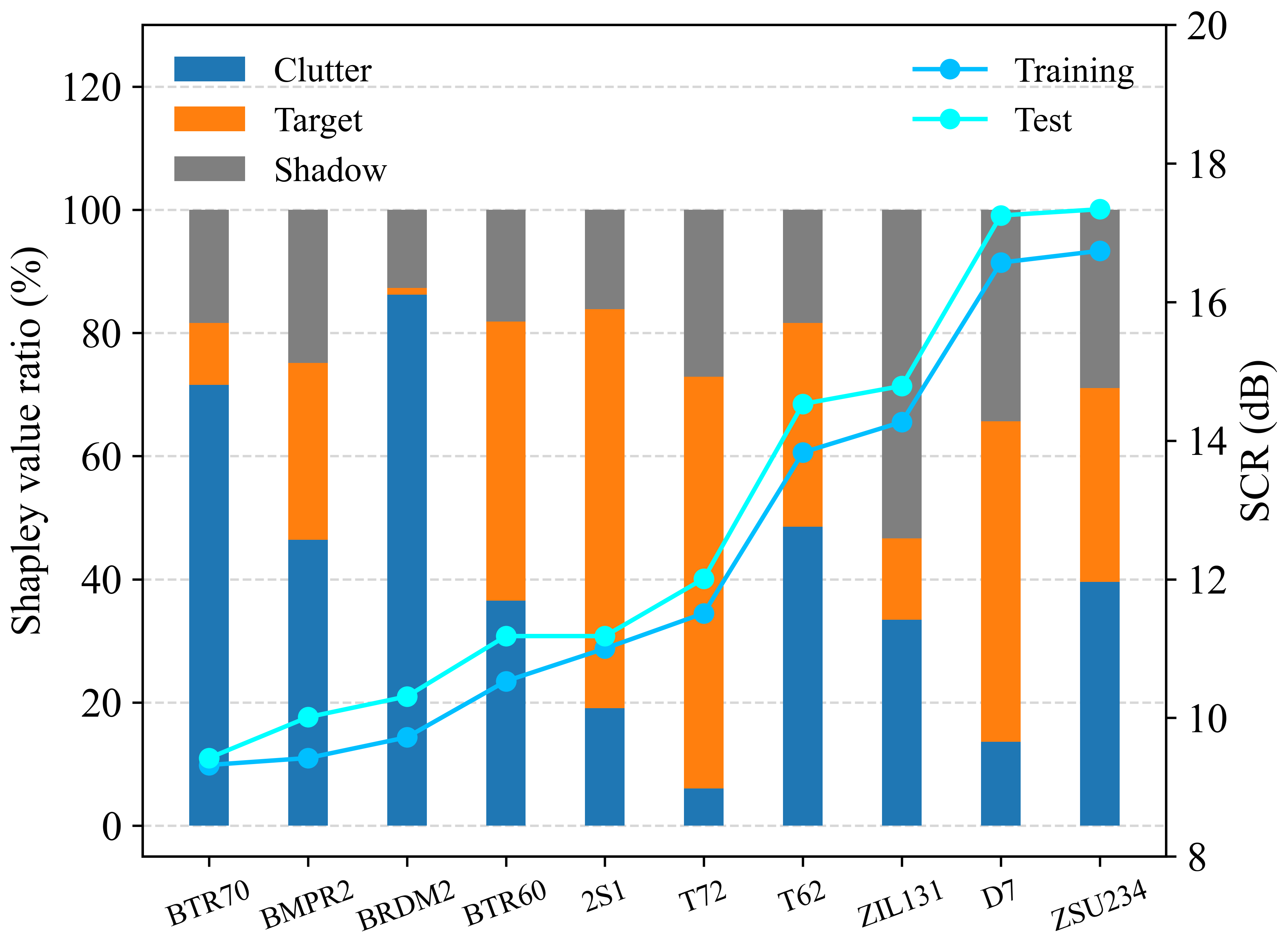}}
\subfloat[ConvNeXtTiny]{\includegraphics[width=4.5cm]{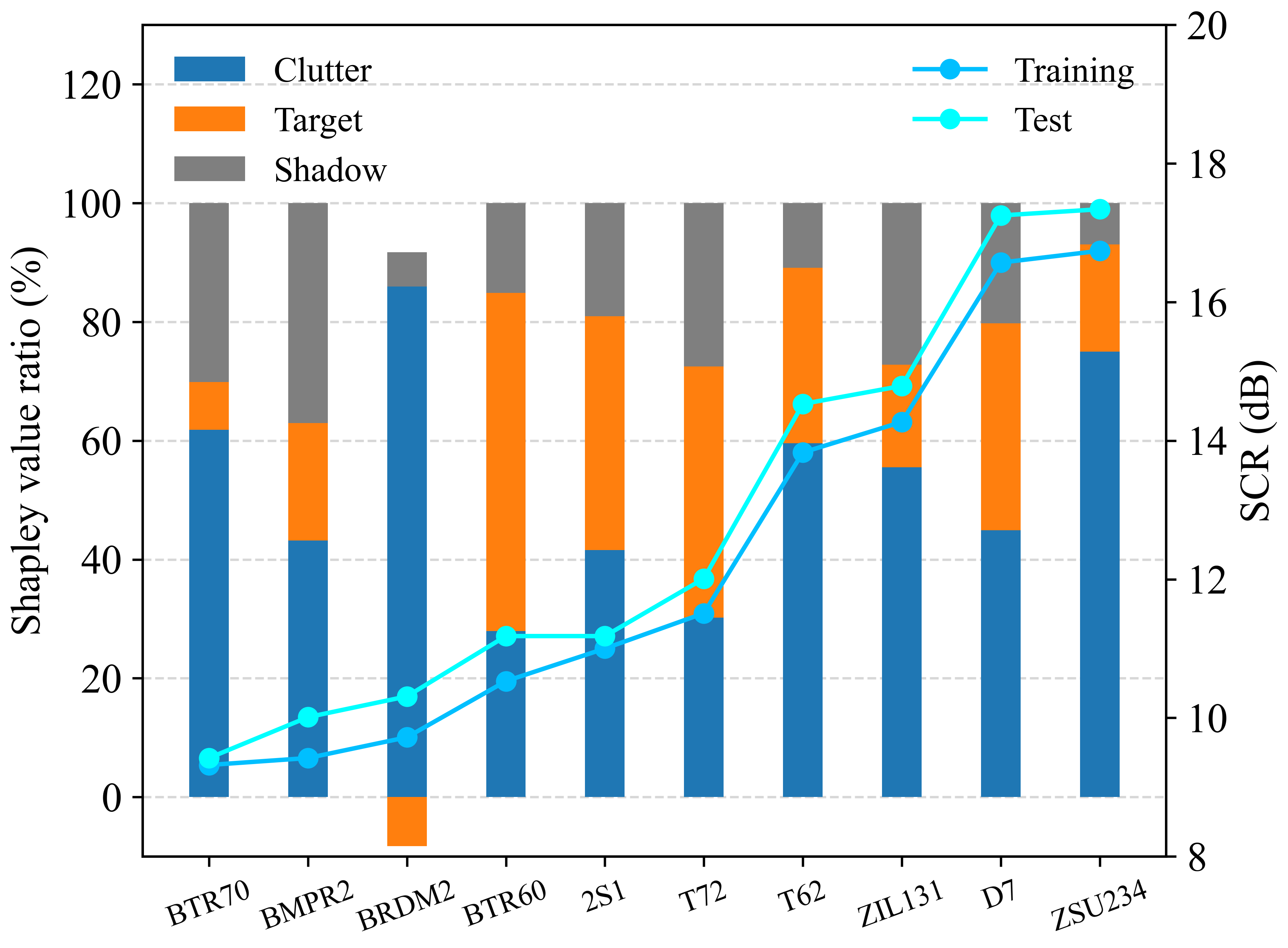}}
\caption{Shapley value ratio and SCR curve across target classes. The first-row models tend to use clutter with low SCR on the left side. The second-row models with optical pre-training weights overfit clutter textures of different classes in addition to low SCR. Many models extract different discriminative features, but these features use clutter incorrectly. In addition, the comparable SCR curve between training and test sets illustrates the spurious correlation of clutter. (A negative Shapley value indicates the adverse effect, and replacing the region with the baseline value helps to recognize this class.)}
\label{fig_class}
\end{figure*}

\textbf{Exp. 1, deep learning uses all correlations to reduce training errors, which does not guarantee causality.} The recognition rate and Shapley values increase simultaneously during training in Fig. \ref{model_1_shapley}. However, the Shapley value ratio shows different trends. The clutter curve undergoes an early rising process in Fig. \ref{model_1_shapley_ratio}. This curve is because deep learning does not distinguish between correlations that come from causality and data bias. A-ConvNet initially focuses more on background differences. According to Table \ref{table0}, shadow areas have the least influence on recognition, with clutter and targets having the most influence. This finding is similar to the previous study\cite{belloni2020explainability}. In summary, the overfitting for clutter reflects the non-causality of deep learning.

\textbf{Exp. 2, deep learning further exploits interactions between regions to reduce training errors.} Fig. \ref{model_1_bshapley} shows that interactions between regions increase during training, indicating that the model can exploit correlations on a large scale. Both the target and shadow regions contain class information, but the model also uses clutter differences to form a coalition. Table \ref{table0} shows that the strength of coalition interaction varies across models. Hence, the interactions of regions in deep learning are also influenced by biases.

\subsection{Explaining the Non-Causality}

\textbf{Exp. 3, data bias leads to class-related clutter properties, one of which is the comparable SCR curves between the training and test sets.} 
According to the SCR curves in Fig. \ref{fig_class}, the background clutter is comparable for different target classes in MSTAR. Due to background correlation in training and test sets, overfitting for clutter does not significantly reduce performance under SOC in Table \ref{table0}. Interestingly, the target region contributes negatively to the results when A-ConvNet recognizes the four lowest SCR classes in Fig \ref{fig_class}. This phenomenon indicates that the model learns incorrect feature representations to recognize some classes. Similarly, removing targets does not affect the recognition rate for some classes in \cite{belloni2020explainability}. Another similar case in computer vision is using snowy backgrounds to distinguish between wolves and huskies\cite{fjelland2020general}. Despite the impressive performance in Table \ref{table0}, our work illustrates that background correlation results in deep learning-based feature representations that are incorrect but valid for biased datasets.  

\begin{figure*}[!tb]
\centering
\includegraphics[width=16cm]{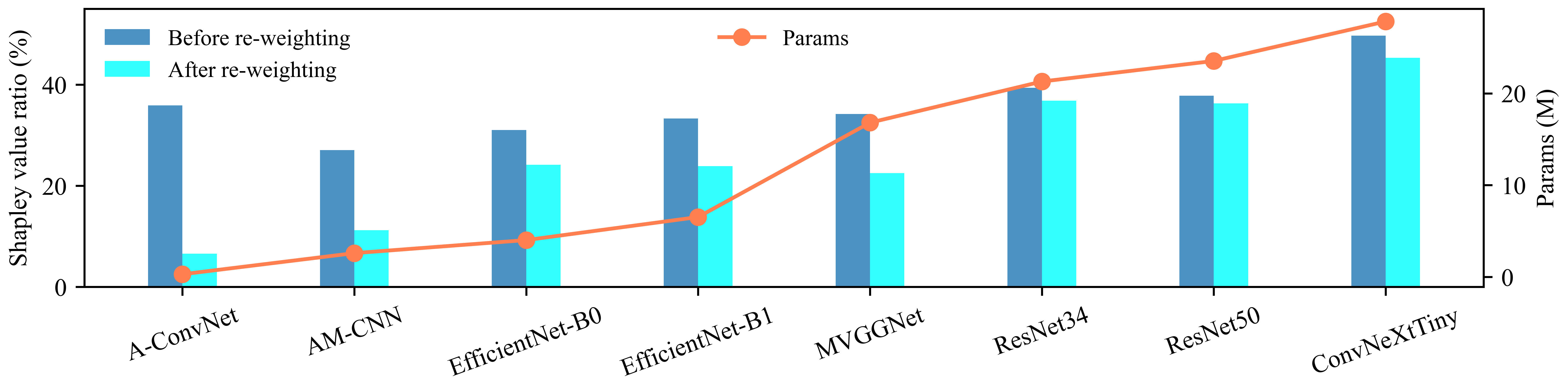}
\caption{Shapley value ratio of clutter with random SCR re-weighting. Although eliminating comparable SCR can reduce clutter contributions, models with a large number of parameters still overfit the background clutter after SCR re-weighting.}
\label{scr_re}
\end{figure*}

\textbf{Exp. 4, deep learning models can use background clutter texture for recognition.} Fig. \ref{scr_re} shows that clutter contribution is reduced by eliminating the comparable SCR. This result also confirms that comparable SCR is one of the factors contributing to background correlation. As our method only changes the amplitude of clutter, the texture remains intact\footnote{The relative intensity of clutter scattering points still maintains the original spatial distribution, which keeps a local repeating clutter texture in SAR images.}. Our experimental results suggest that the model can more memorize clutter texture in datasets when its capacity increases significantly with the parameter\footnote{We use parameters to measure different models. In fact, many factors influence texture bias. We find that optimization methods and model structures affect clutter contribution. Therefore, we use the same optimizer. Furthermore, special model structures such as batch normalization and attentional mechanisms need to be discussed to investigate whether the texture bias is discrepant for the target or clutter.}.

\section{Conclusion}
We analyzed the non-causality of deep learning in SAR ATR based on the Shapley value. Data and model biases lead to non-causality and overfitting for background clutter in deep learning. Background correlation leads to comparable SCR and clutter texture in MSTAR, and the models with different parameters have different degrees of overfitting to these clutter properties. In addition, overfitting for clutter is normally hidden by the background correlation of the MSTAR dataset and the black-box property of deep learning. Our analysis indicates that the causality and robustness of deep learning in SAR ATR are still under consideration. Since clutter is color noise, overfitting for clutter implies non-robustness in various environments. Moreover, causal features (\emph{i.e.}, target signatures) are unstable under different imaging conditions. Current adversarial attack studies have shown that deep learning is affected by small shifts in target signatures\cite{9800917,9915465}. A small dataset such as MSTAR can not adequately reflect target and background variations across imaging conditions, even with reduced data bias. It remains to be explored whether deep learning can achieve robust SAR ATR. Deep learning requires causality to suppress clutter and be robust to shifts in target signatures for real-world applications.

\bibliographystyle{IEEEtran}
\bibliography{ref}

\end{document}


\title{Discovering and Explaining the Non-Causality of Deep Learning in SAR ATR: Supplementary Material}
\markboth{In preparation for submission}%
{Li \MakeLowercase{\textit{et onclusial.}}: A Sample Article Using IEEEtran.cls for IEEE Journals}

\maketitle

\subsection{The Training of Models}
\begin{figure*}[!htb]
\centering
\subfloat[A-ConvNet]{\includegraphics[width=4.5cm]{./Pic/Model_1_shapley.png}}
\subfloat[AM-CNN]{\includegraphics[width=4.5cm]{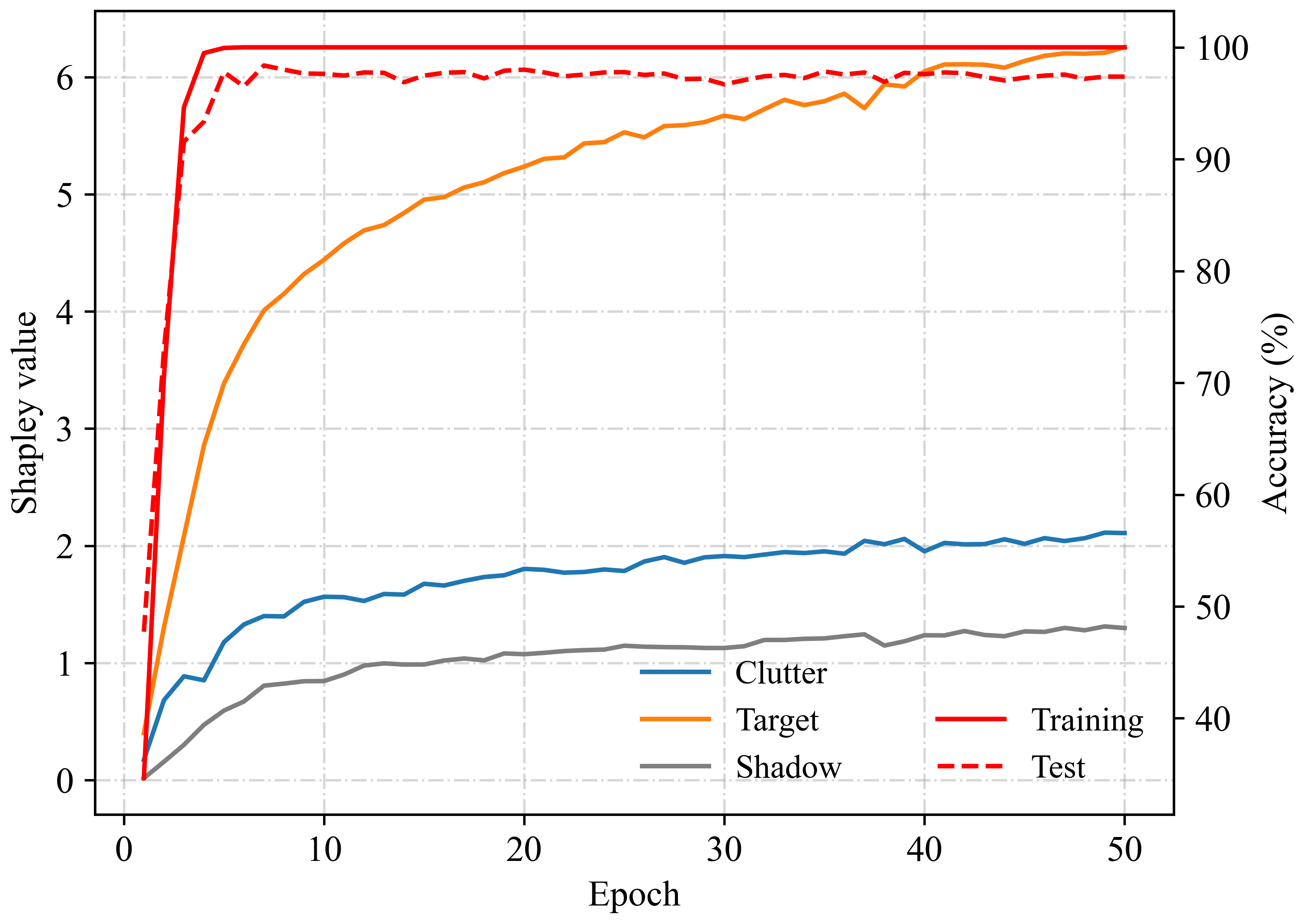}}
\subfloat[EfficientNet-B0]{\includegraphics[width=4.5cm]{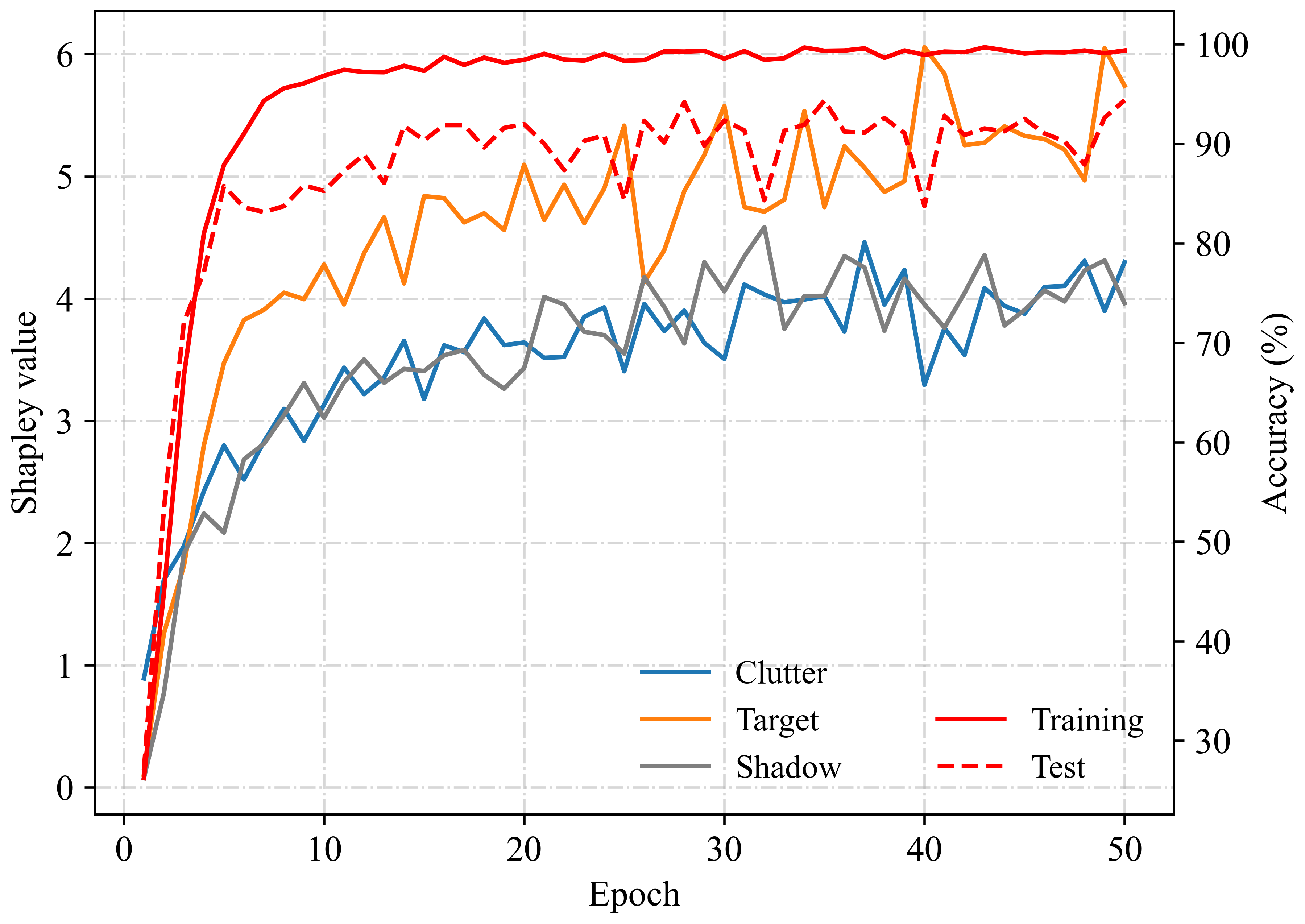}}
\subfloat[EfficientNet-B1]{\includegraphics[width=4.5cm]{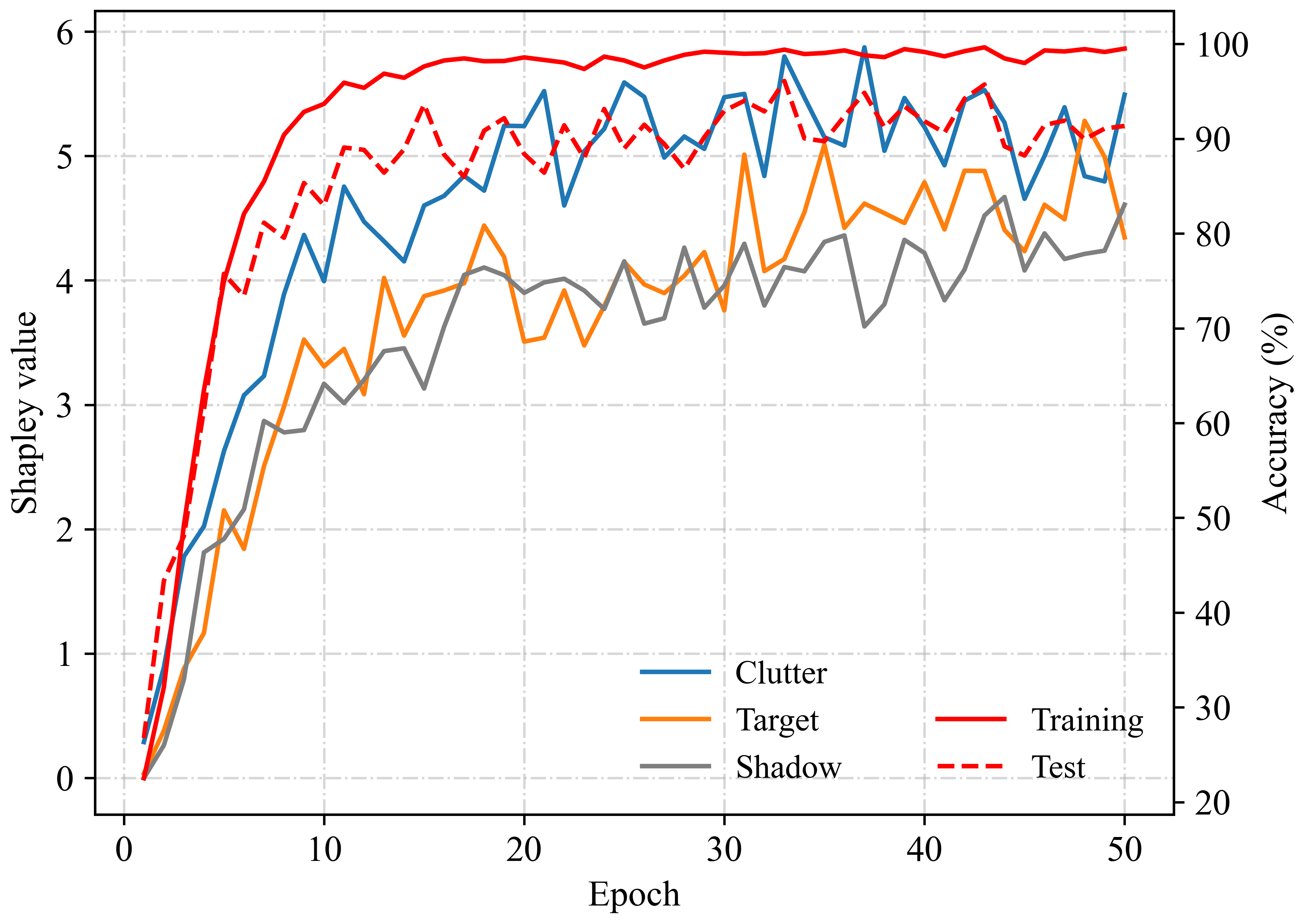}}
\vfil
\subfloat[MVGGNet]{\includegraphics[width=4.5cm]{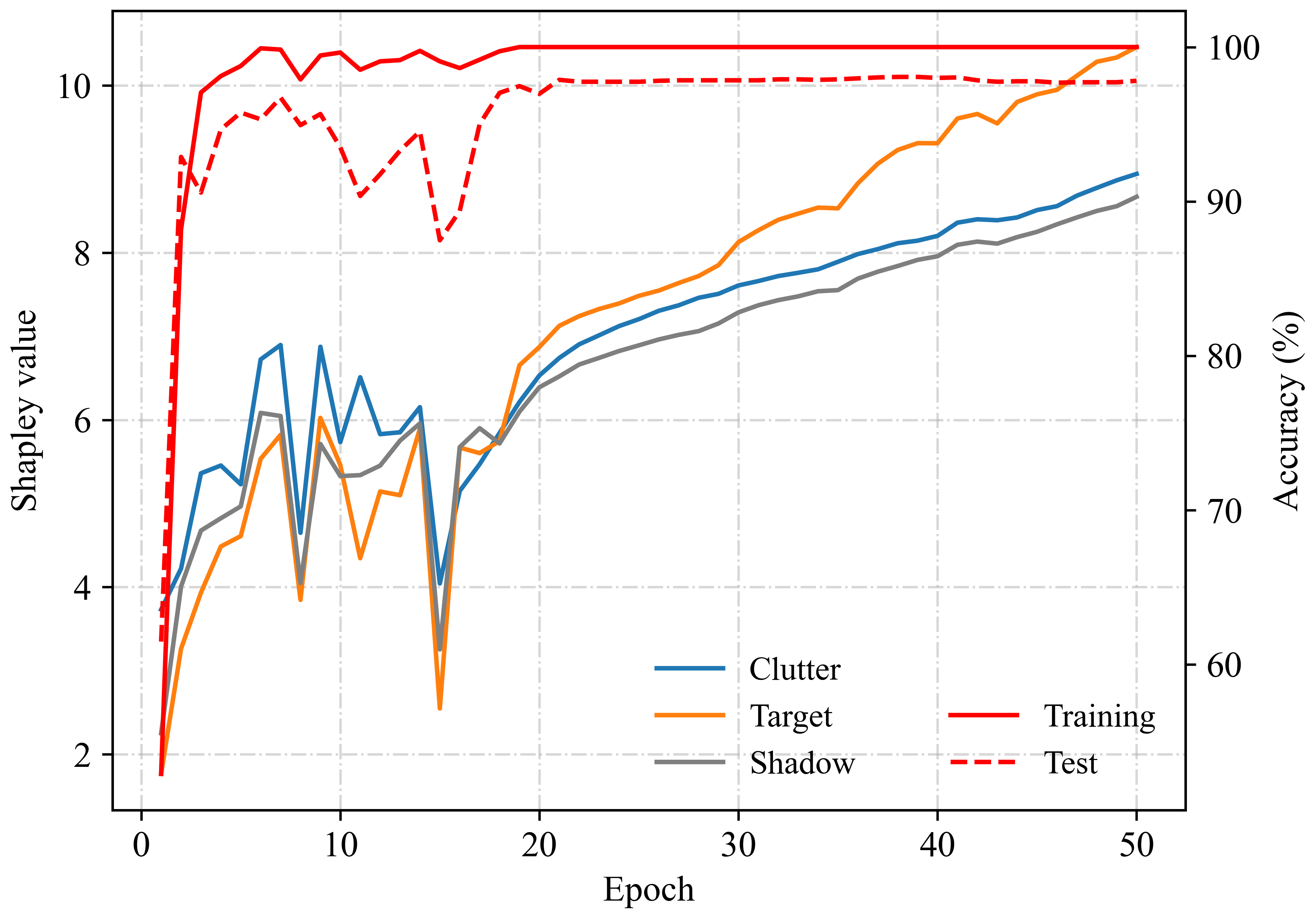}}
\subfloat[ResNet34]{\includegraphics[width=4.5cm]{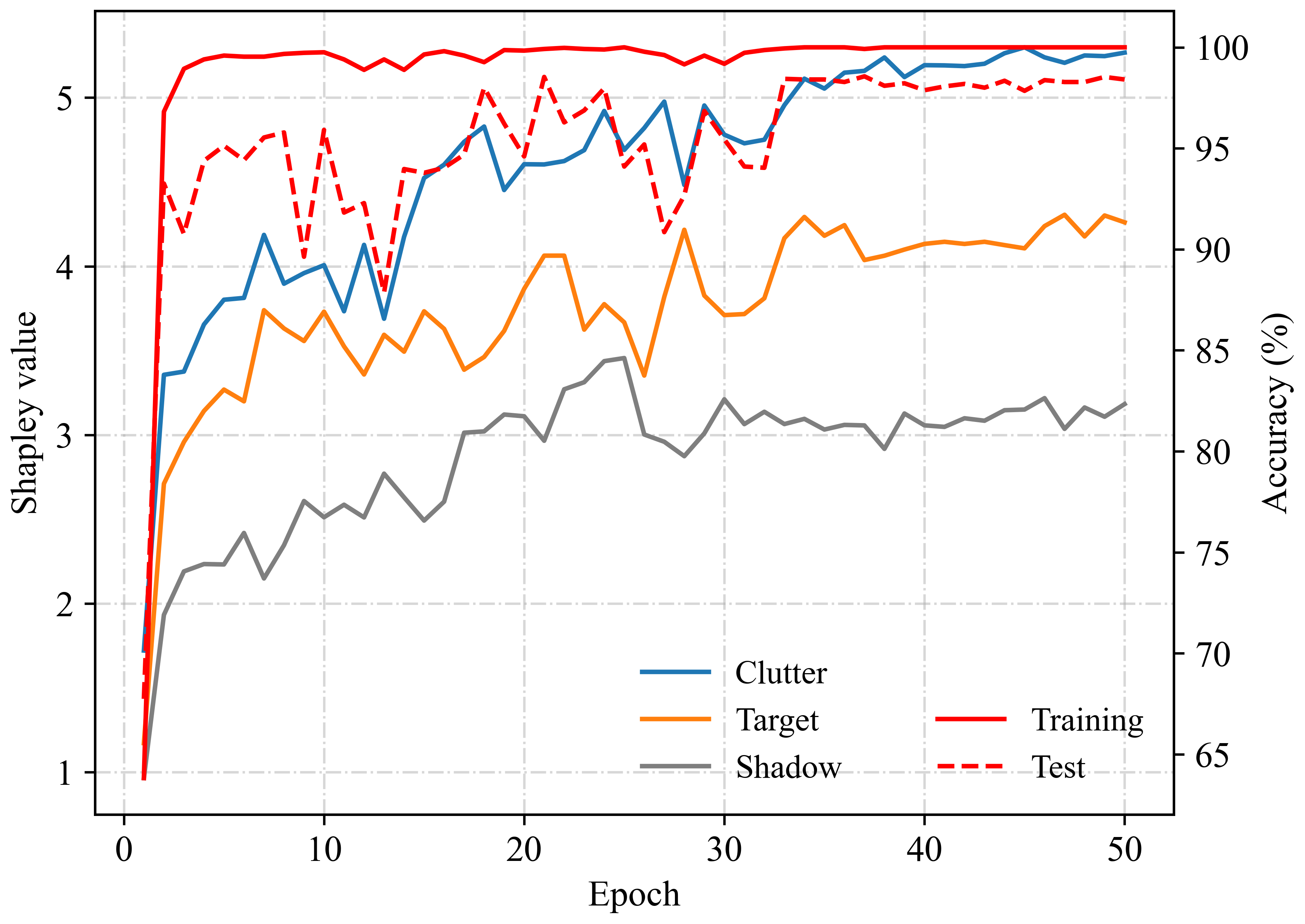}}
\subfloat[ResNet50]{\includegraphics[width=4.5cm]{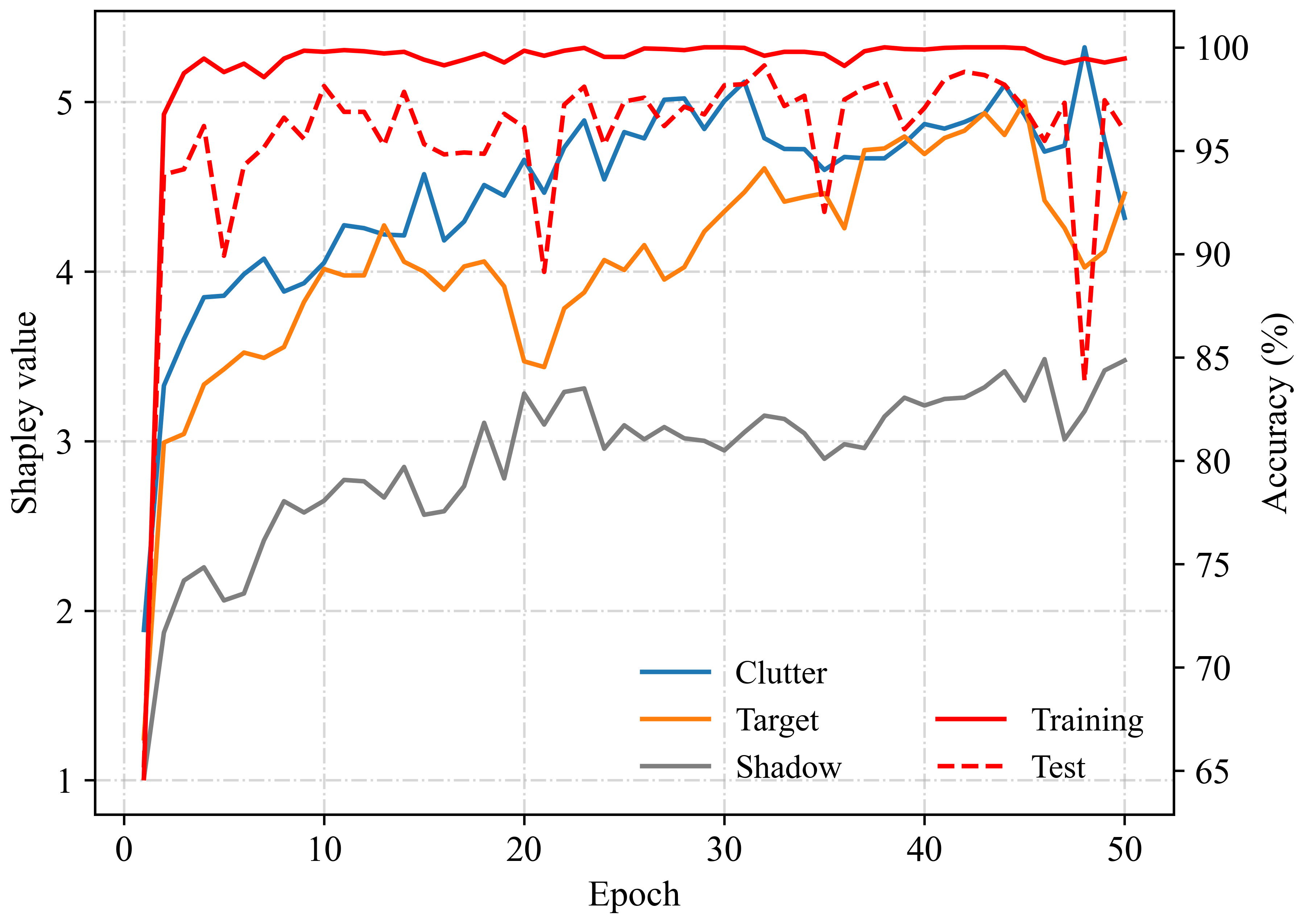}}
\subfloat[ConvNeXtTiny]{\includegraphics[width=4.5cm]{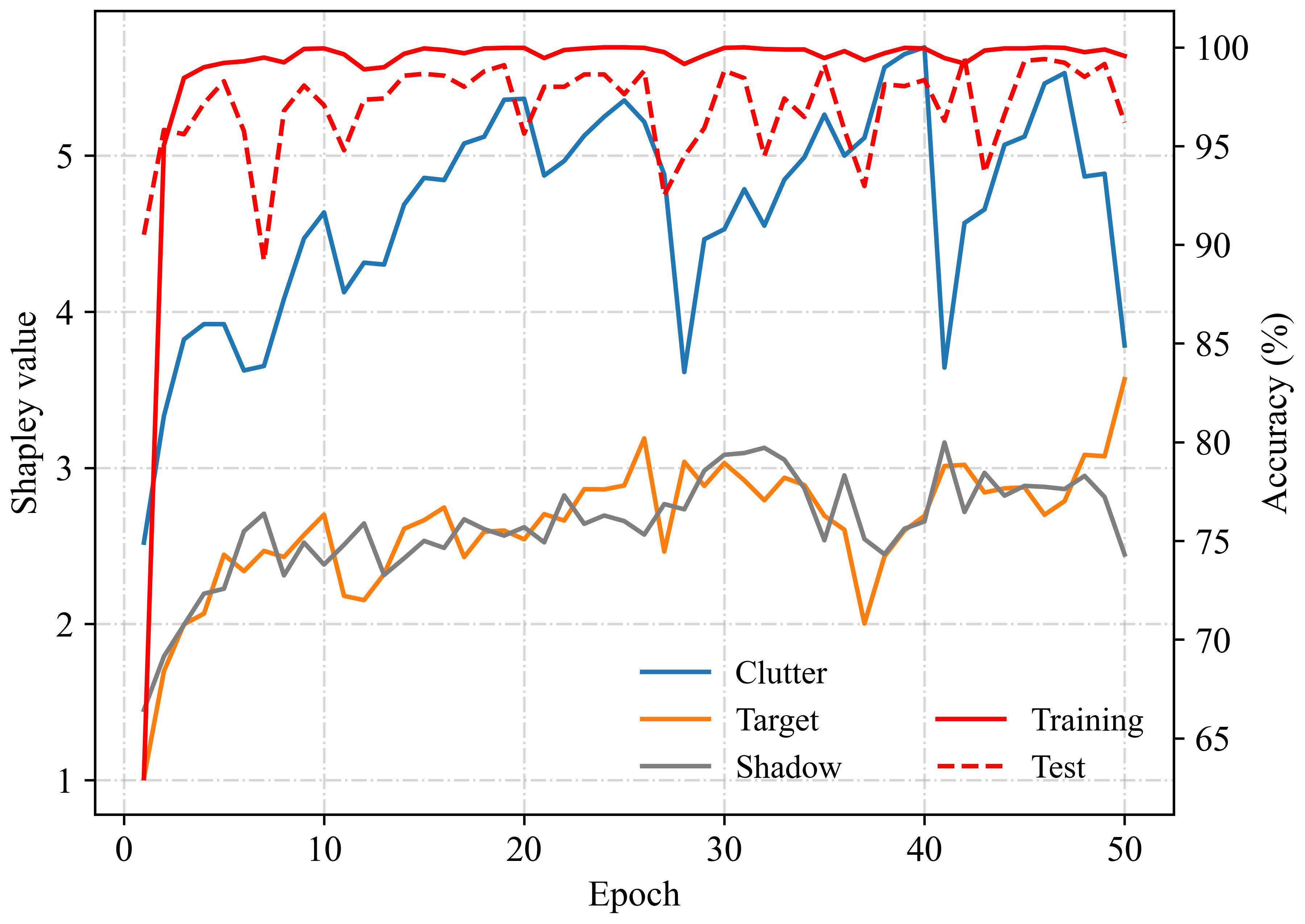}}
\caption{Shapley value during the training.}
\label{1}
\end{figure*}

\begin{figure*}[!htb]
\centering
\subfloat[A-ConvNet]{\includegraphics[width=4.5cm]{./Pic/Model_1_shapley_ratio.png}}
\subfloat[AM-CNN]{\includegraphics[width=4.5cm]{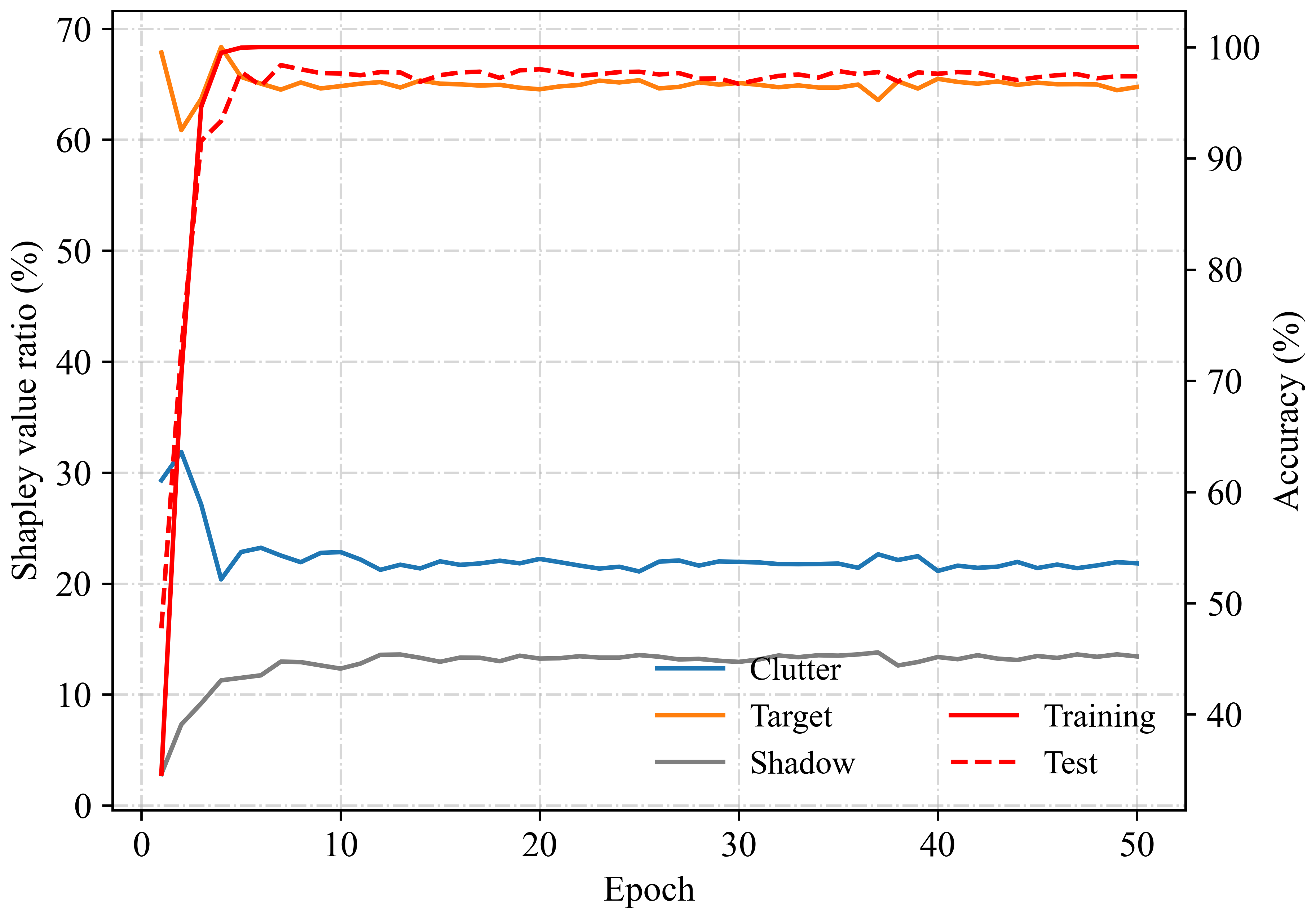}}
\subfloat[EfficientNet-B0]{\includegraphics[width=4.5cm]{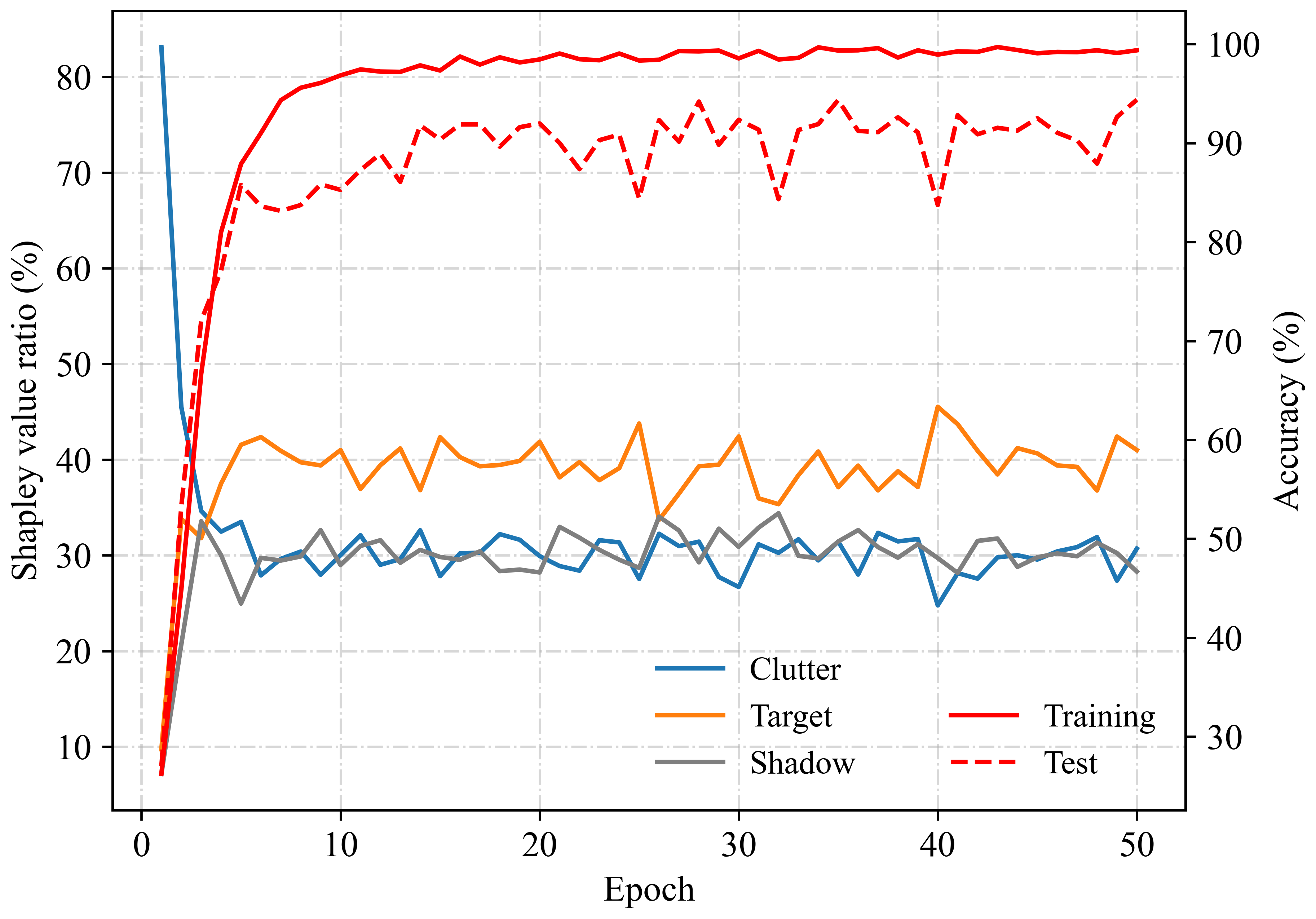}}
\subfloat[EfficientNet-B1]{\includegraphics[width=4.5cm]{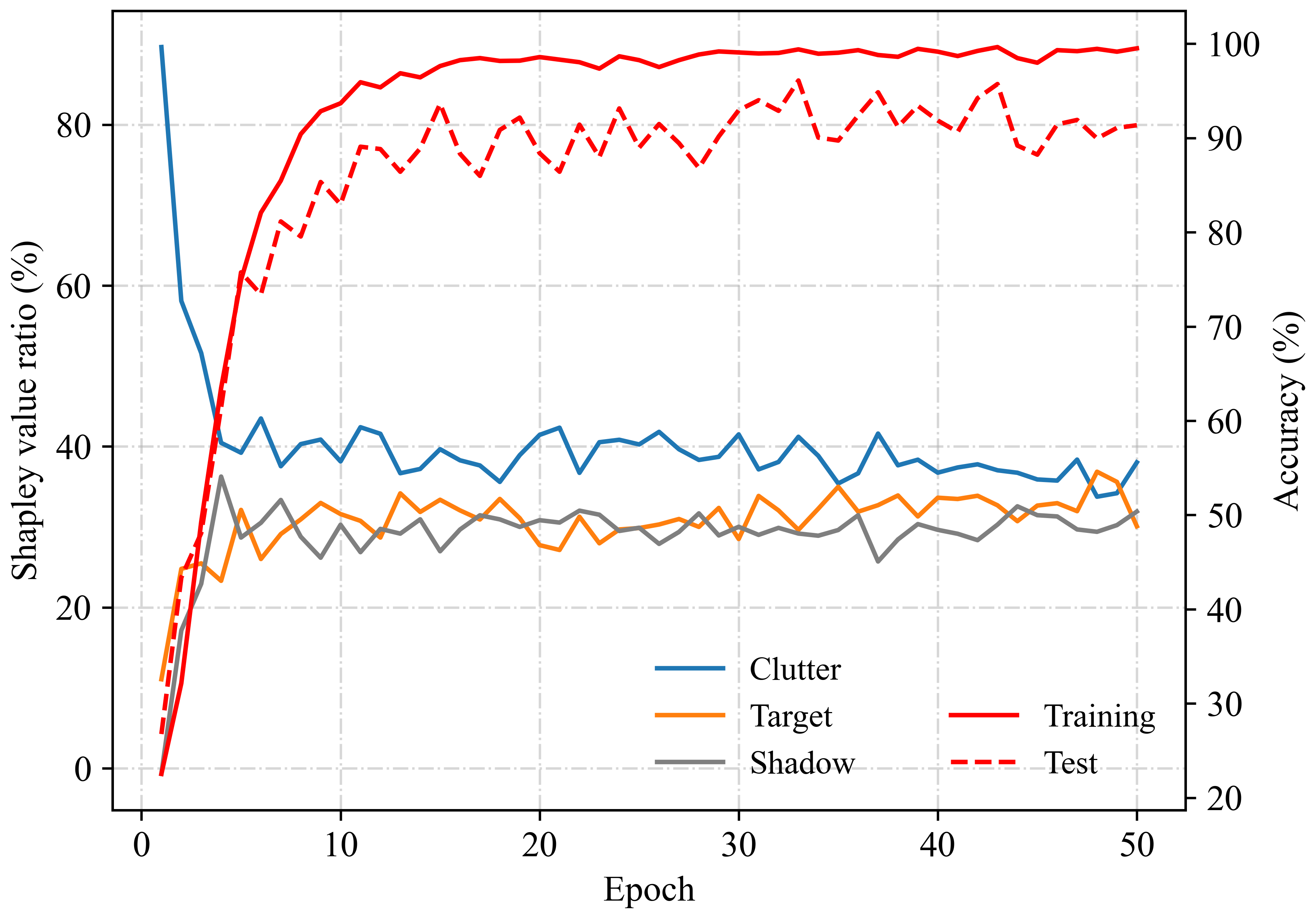}}
\vfil
\subfloat[MVGGNet]{\includegraphics[width=4.5cm]{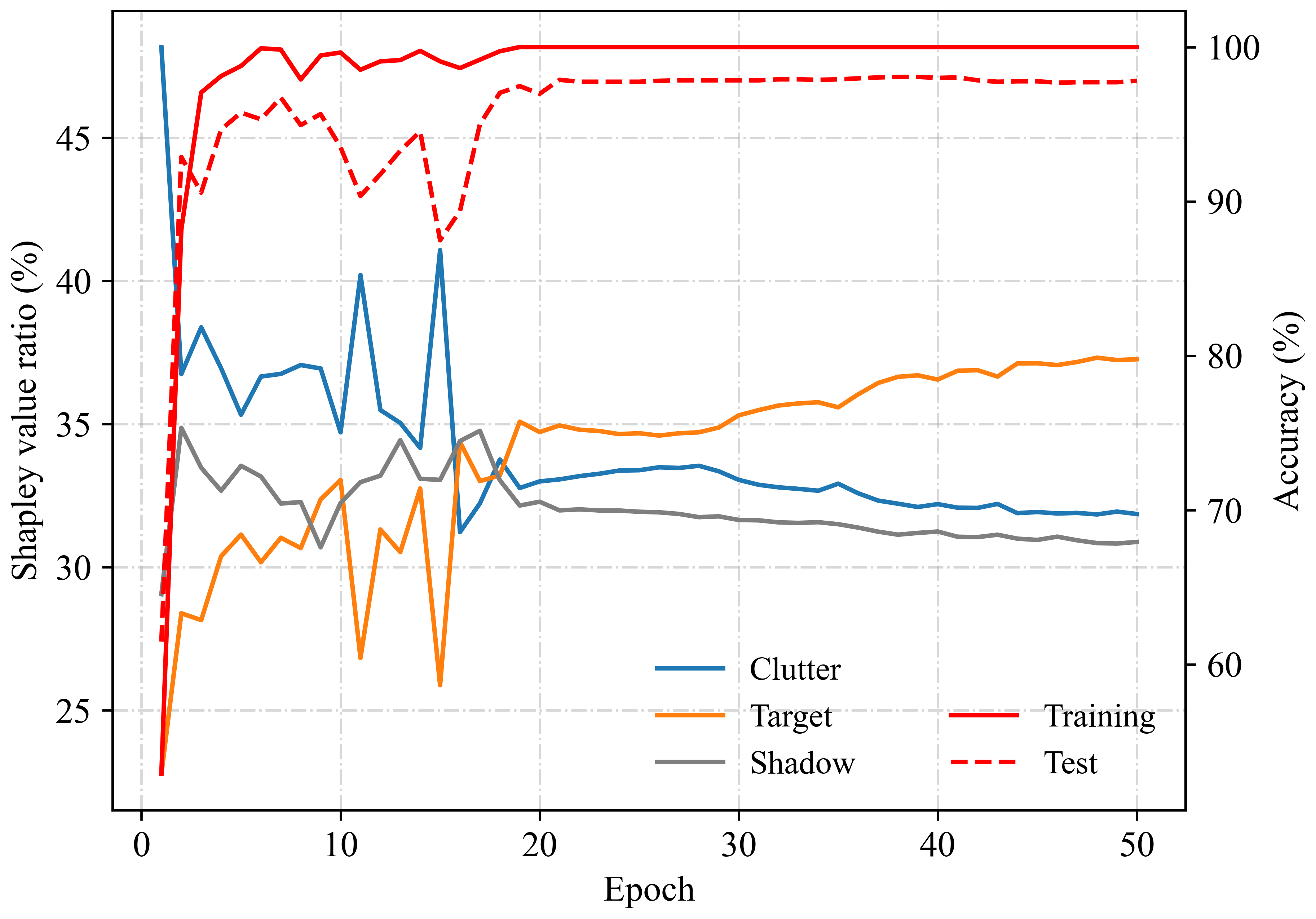}}
\subfloat[ResNet34]{\includegraphics[width=4.5cm]{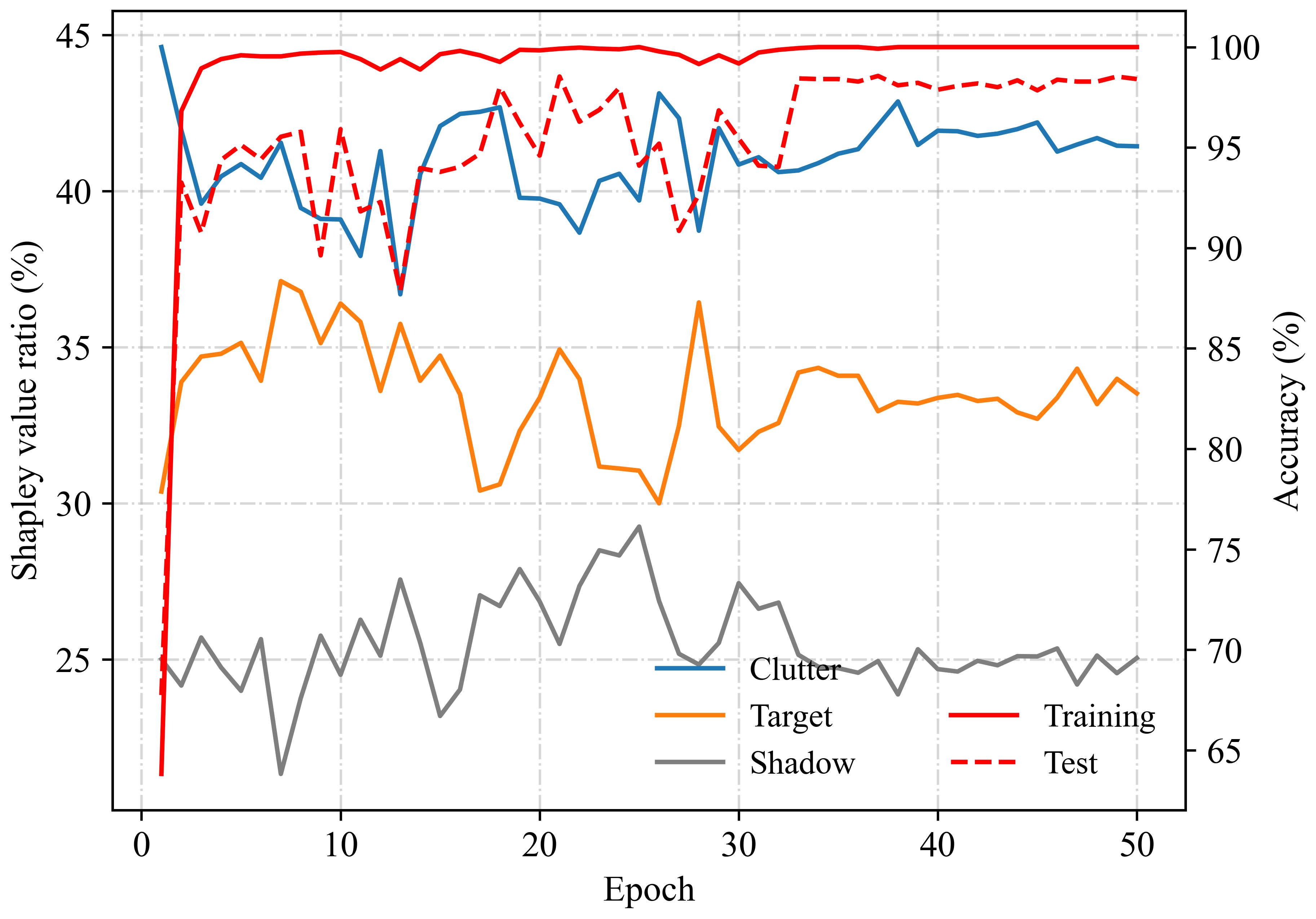}}
\subfloat[ResNet50]{\includegraphics[width=4.5cm]{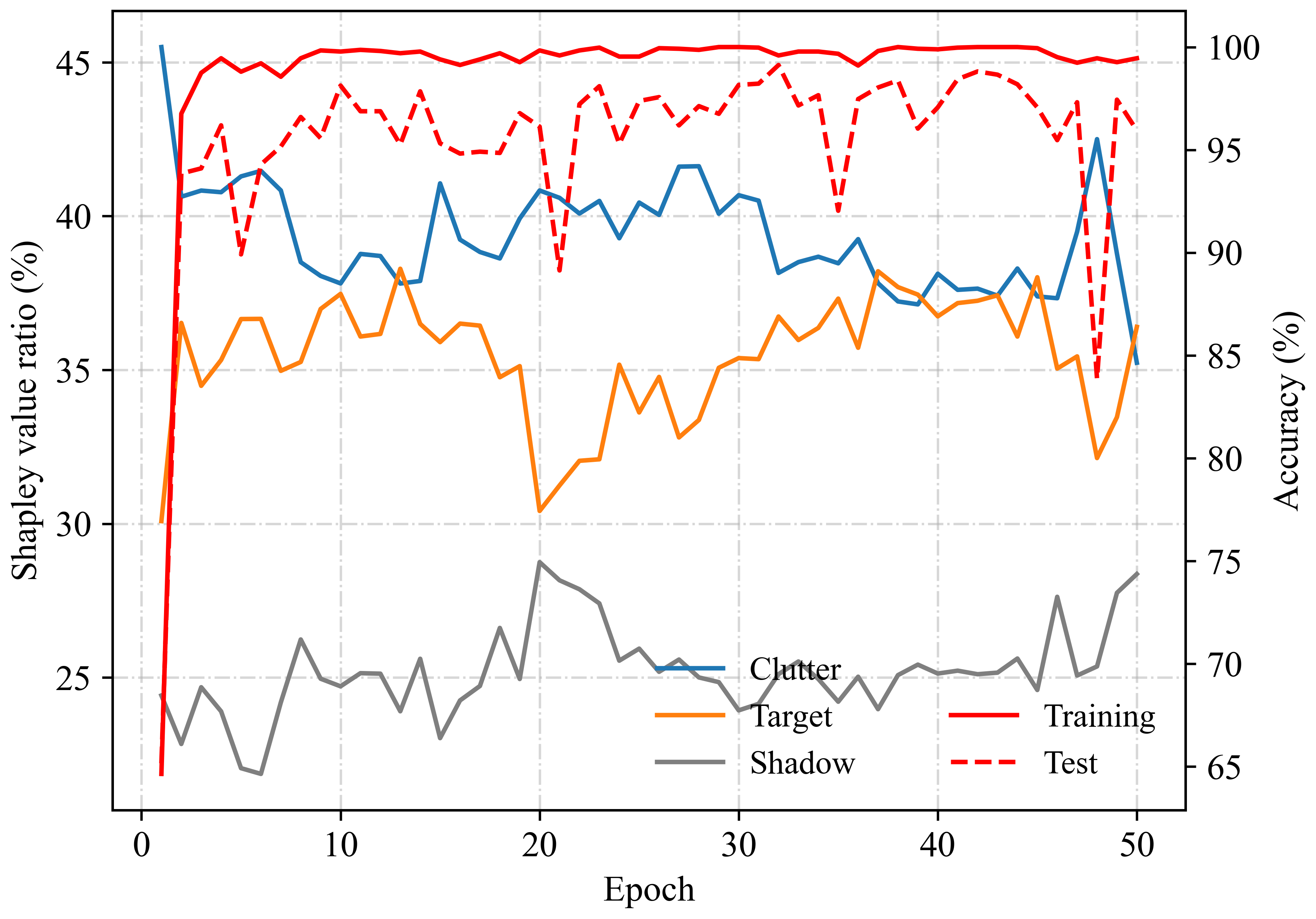}}
\subfloat[ConvNeXtTiny]{\includegraphics[width=4.5cm]{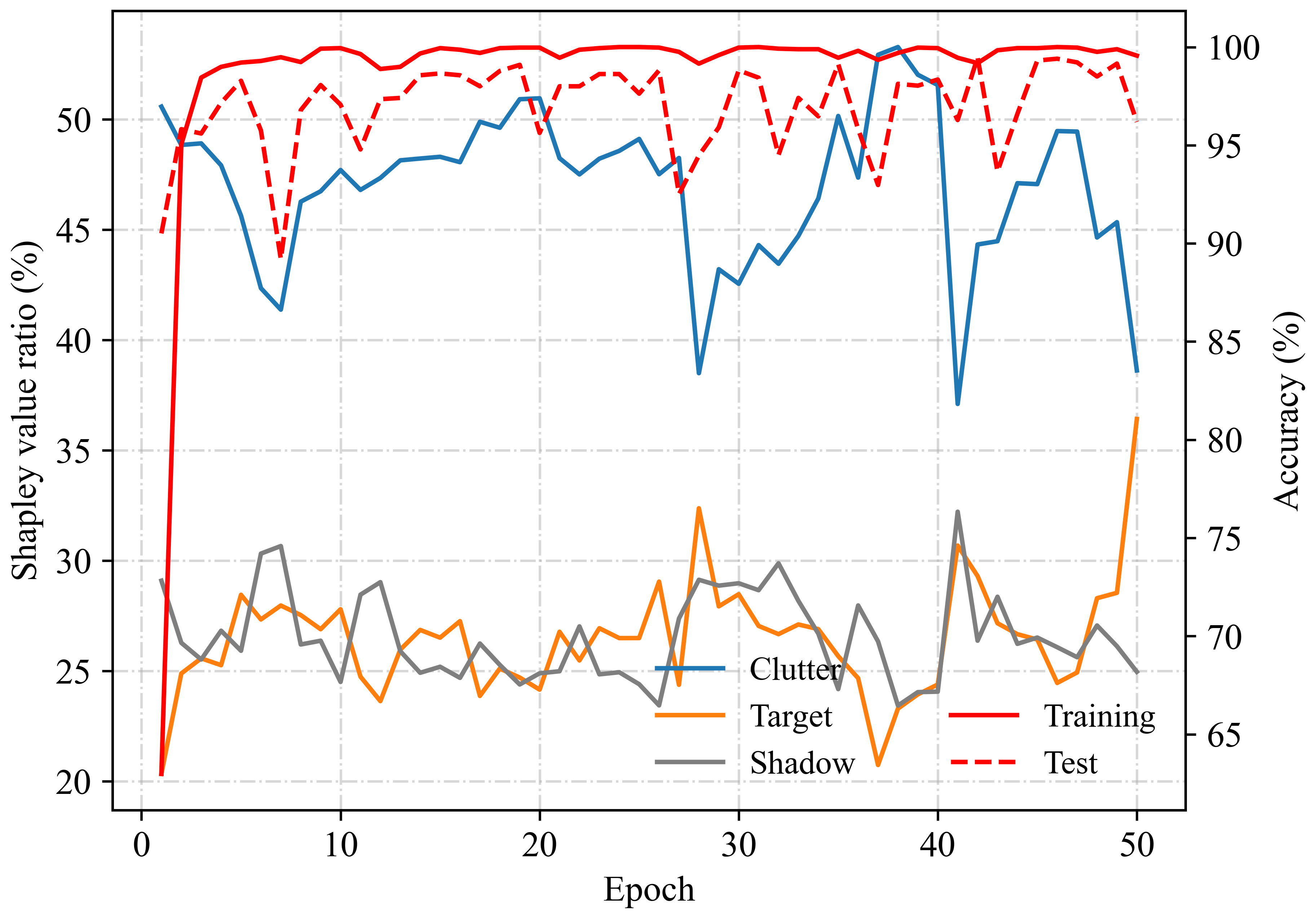}}
\caption{Shapley value ratio during the training.}
\label{2}
\end{figure*}

\begin{figure*}[!htb]
\centering
\subfloat[A-ConvNet]{\includegraphics[width=4.5cm]{./Pic/Model_1_bshapley.png}}
\subfloat[AM-CNN]{\includegraphics[width=4.5cm]{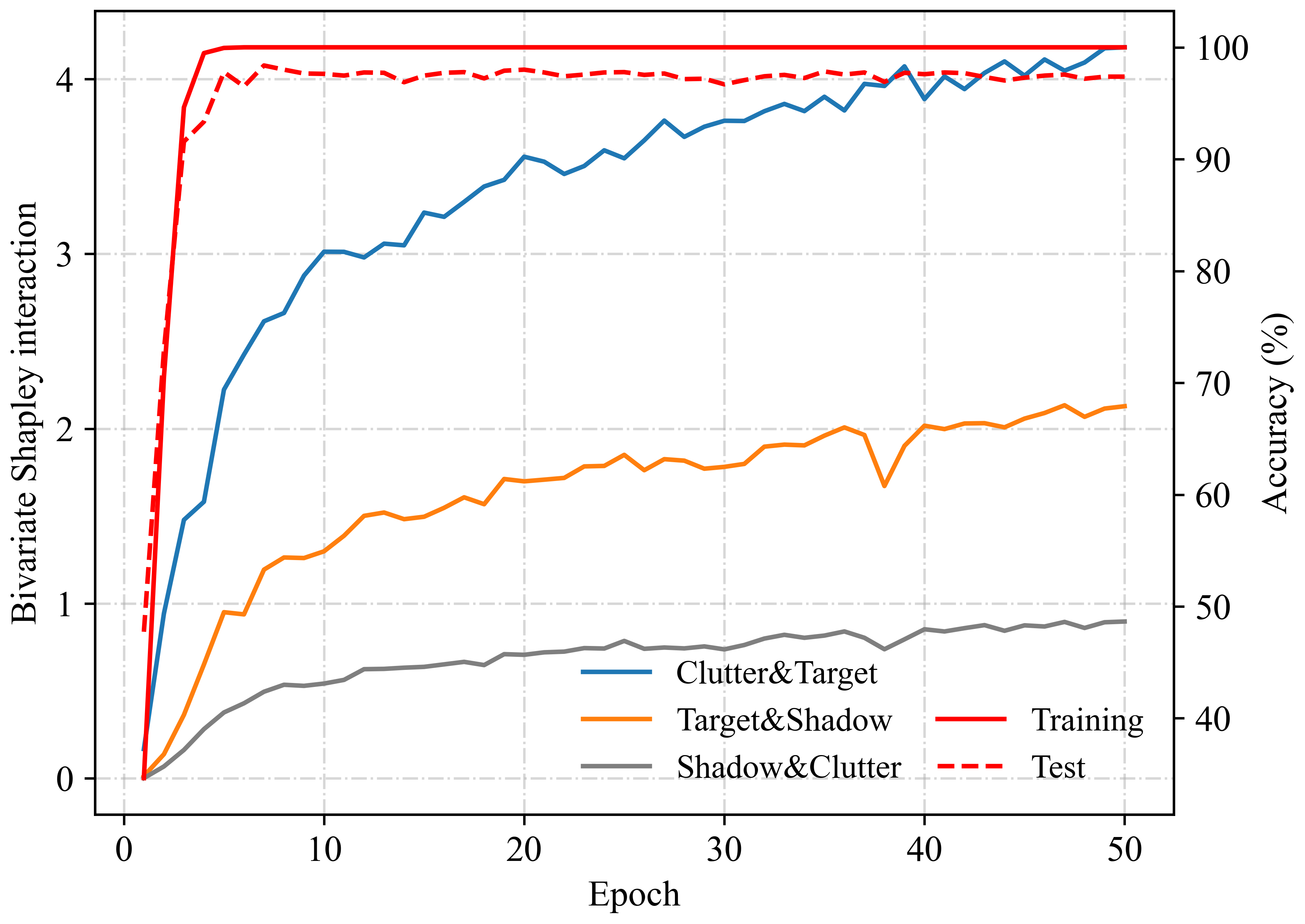}}
\subfloat[EfficientNet-B0]{\includegraphics[width=4.5cm]{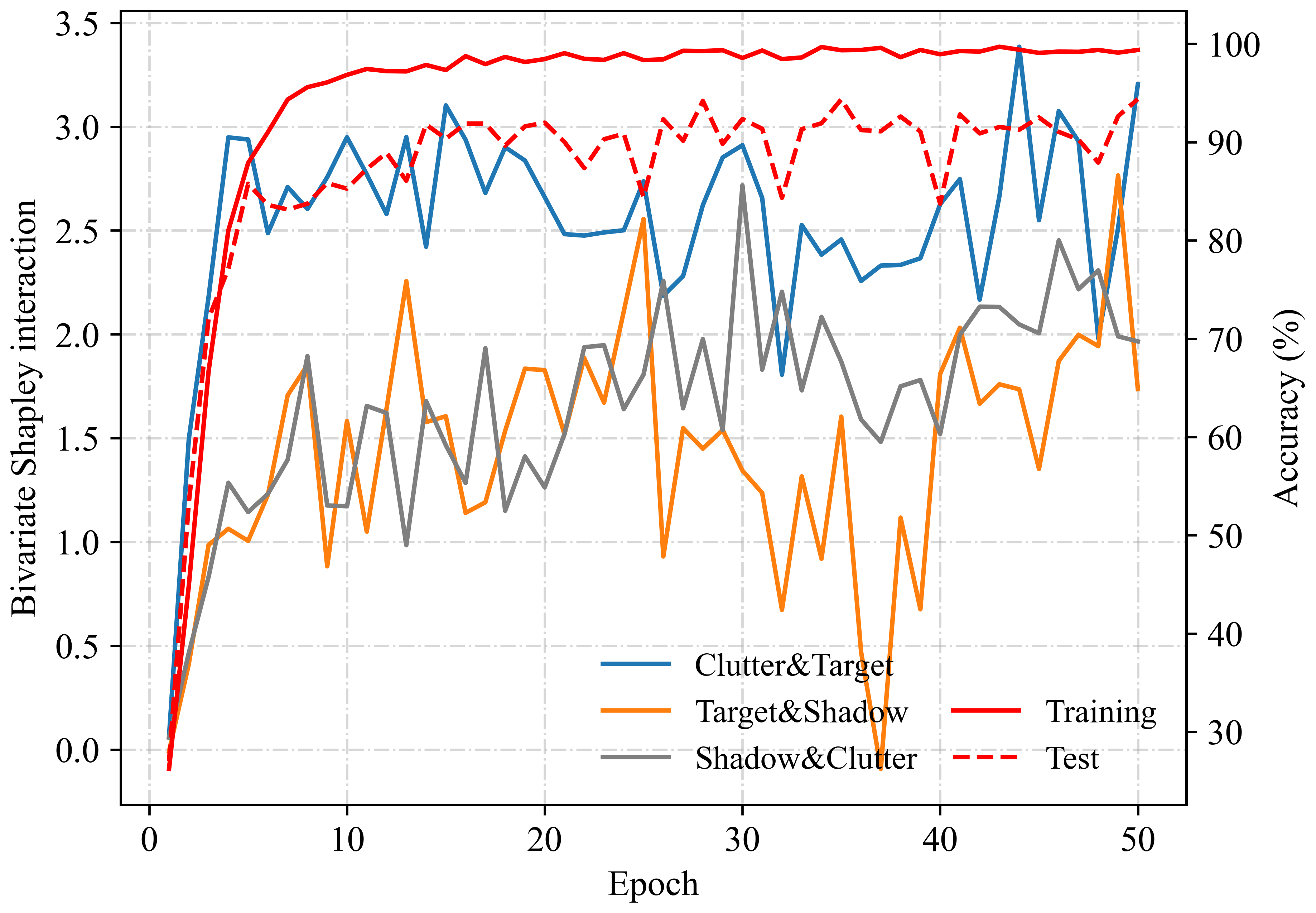}}
\subfloat[EfficientNet-B1]{\includegraphics[width=4.5cm]{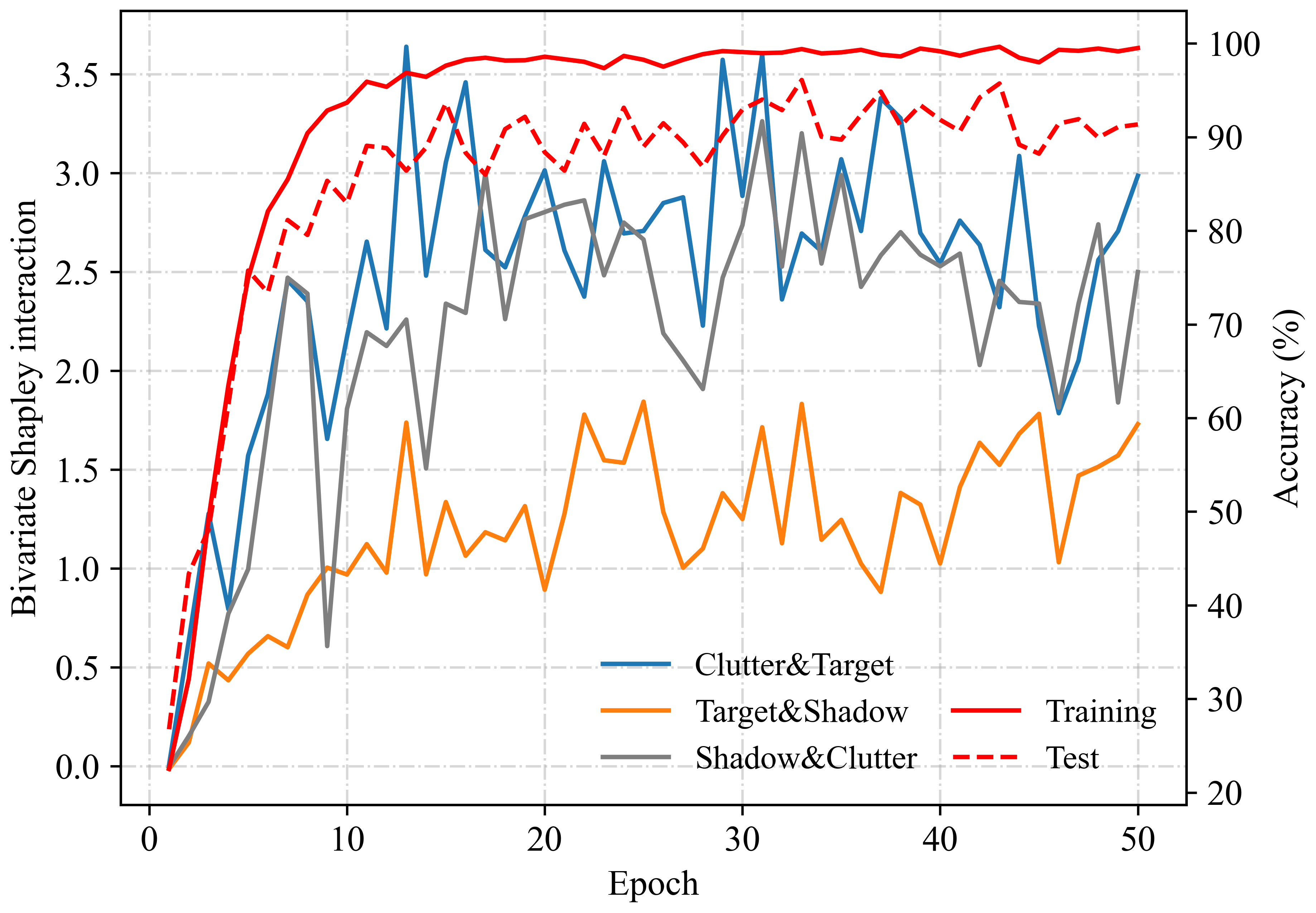}}
\vfil
\subfloat[MVGGNet]{\includegraphics[width=4.5cm]{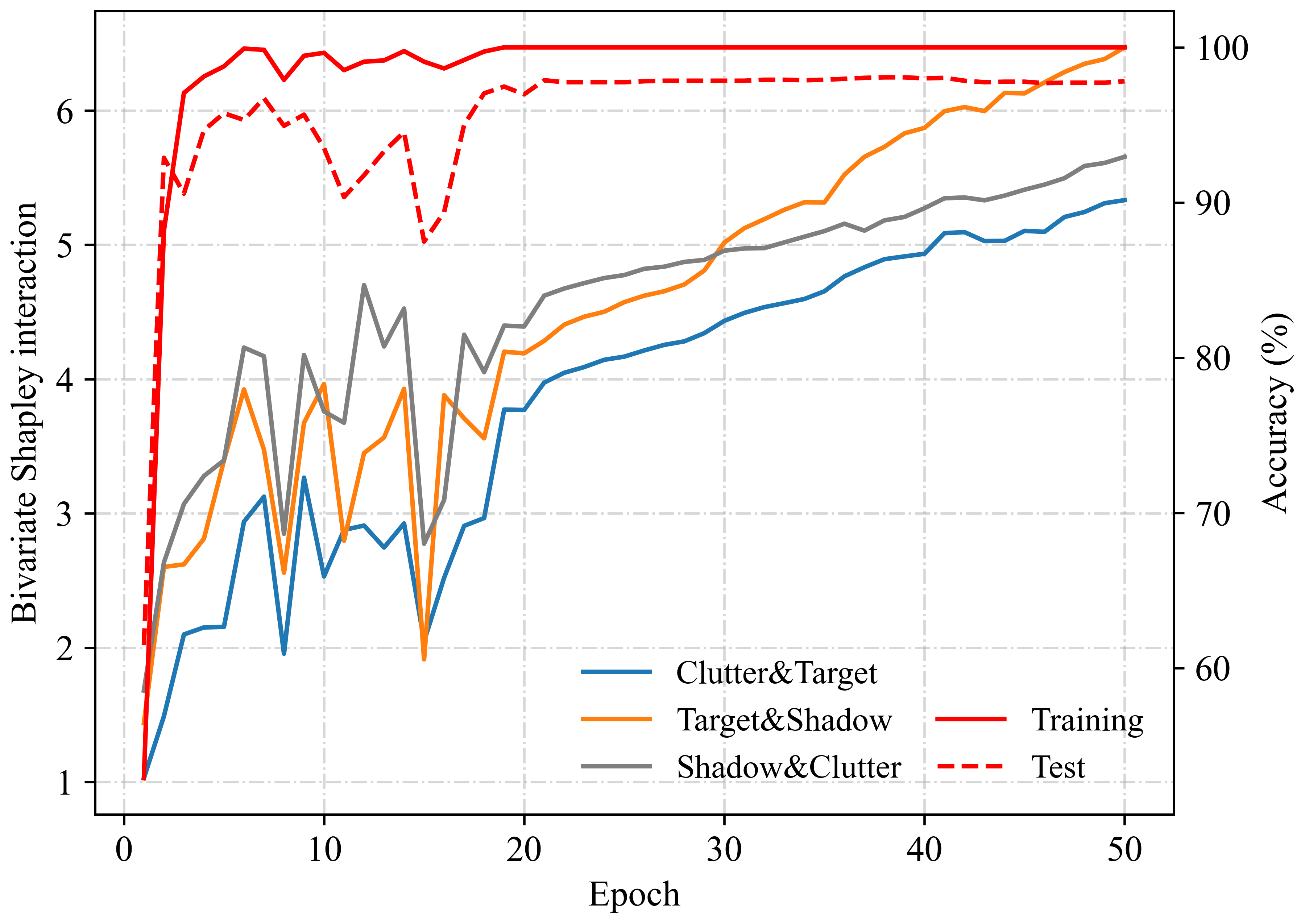}}
\subfloat[ResNet34]{\includegraphics[width=4.5cm]{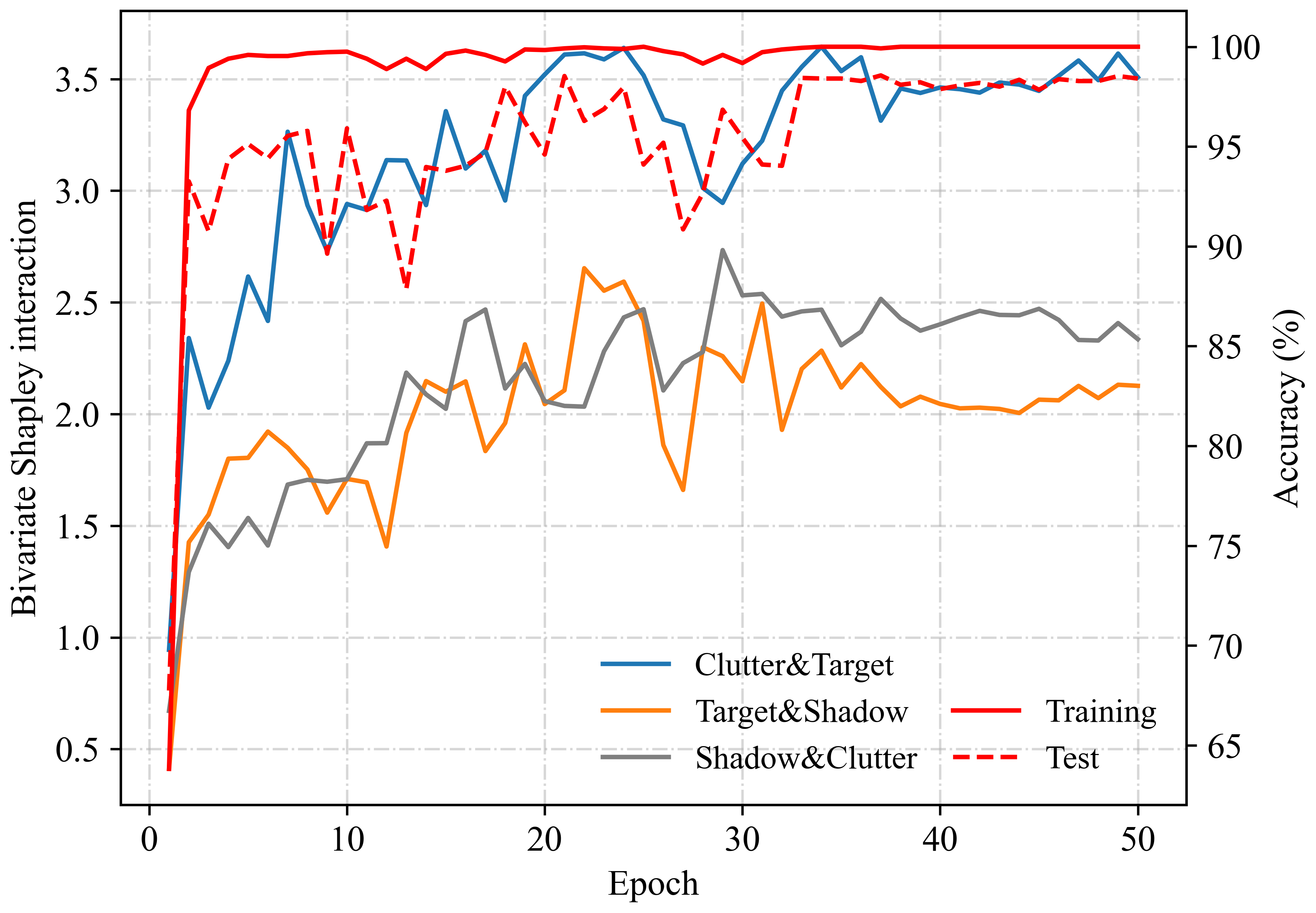}}
\subfloat[ResNet50]{\includegraphics[width=4.5cm]{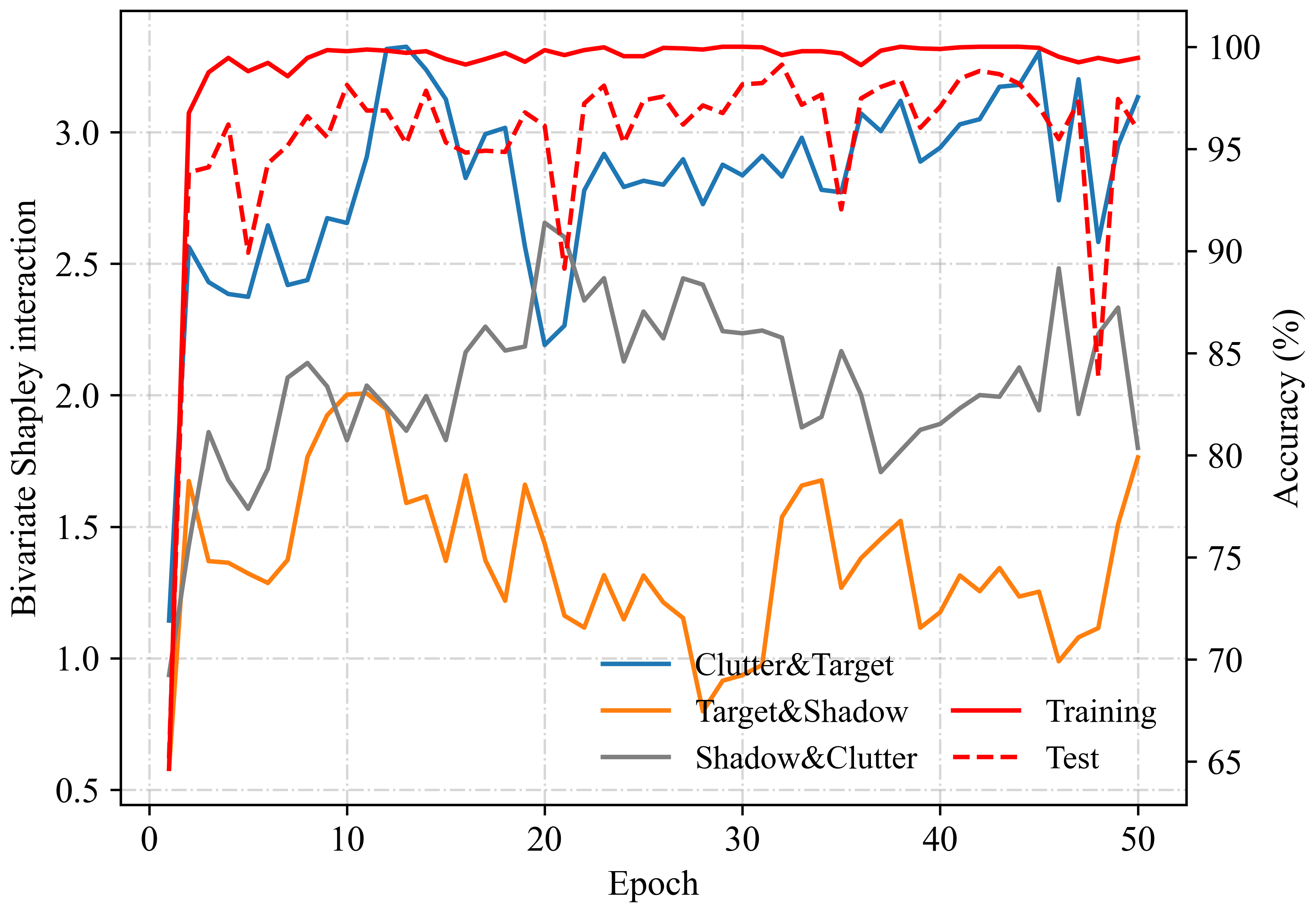}}
\subfloat[ConvNeXtTiny]{\includegraphics[width=4.5cm]{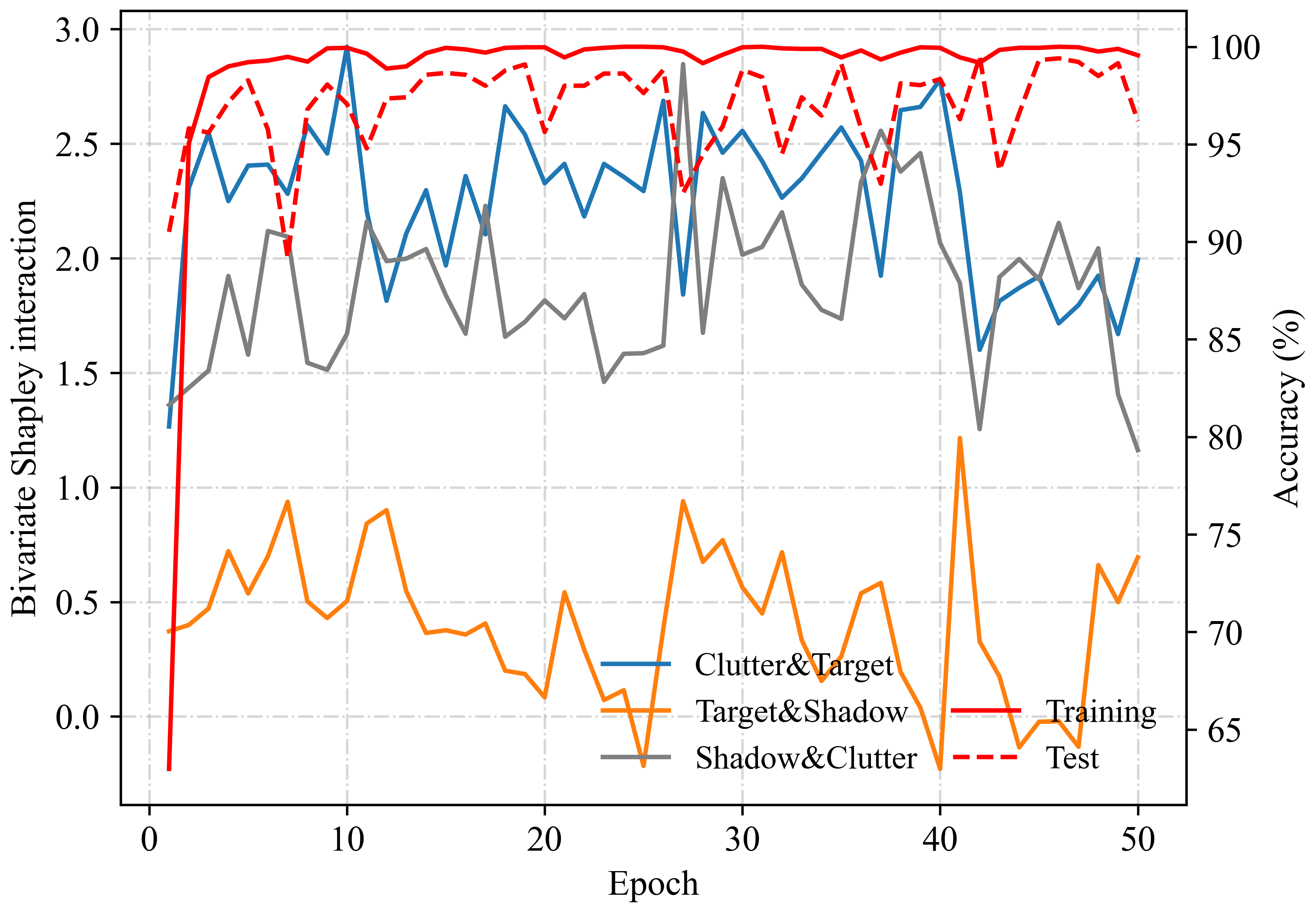}}
\caption{Bivariate Shapley value during the training.}
\label{3}
\end{figure*}

The contributions and interactions of different regions show varying trends across models in Fig. \ref{1}, \ref{2}, and \ref{3}, but the same thing remains: clutter always occupies a certain percentage.

\vspace{1cm}

\subsection{Background Correlation}

\begin{table*}[!htb]
\centering
\caption{Similar Properties of Clutter Between Training and Test Sets}
\label{table1}
\renewcommand\arraystretch{1.35}
\begin{tabular}{cccccccccccc} \toprule
Dataset & Clutter & BTR70 & BMPR2 & BRDM2 & BTR60 & 2S1 & T72 & T62 & ZIL131 & D7 & ZSU234 \\ \midrule
\multirow{2}{*}{Training} & SCR (dB) & 9.32 & 9.42 & 9.72 & 10.53 & 11.00 & 11.51 & 13.83 & 14.27 & 16.57 & 16.74 \\
 & Mean Value & 0.215 & 0.195 & 0.190 & 0.177 & 0.140 & 0.127 & 0.105 & 0.117 & 0.098 & 0.084 \\
\multirow{2}{*}{Test} & SCR (dB) & 9.42 & 10.01 & 10.31 & 11.18 & 11.18 & 12.01 & 14.53 & 14.79 & 17.25 & 17.34 \\
 & Mean Value & 0.211 & 0.184 & 0.176 & 0.164 & 0.130 & 0.122 & 0.095 & 0.107 & 0.090 & 0.077 \\ \bottomrule
\end{tabular}
\end{table*}

\subsection{Result with Random SCR Re-Weighting}
The re-weighting increases the clutter difference between training and test sets, which reduces test set accuracy. These results indicate that deep learning training on the MSTAR small dataset needs to be more robust to background changes. Further analysis of incorrect recognition is needed.

\begin{table*}[!htb]
\centering
\renewcommand\arraystretch{1.35}
\caption{Shapley Value and Bivariate Shapley Interaction of Models with Random SCR Re-Weighting}
\label{table2}
\resizebox{\linewidth}{!}{%
\begin{tabular}{cccccccccc} \toprule
\multirow{2}{*}{Model} & \multicolumn{3}{c}{Region (SVR)} & \multicolumn{3}{c}{Coalition (BSI)} & \multicolumn{2}{c}{Accuracy} & \multirow{2}{*}{Params} \\ \cmidrule(l){2-7}\cmidrule(lr){8-9}
 & Clutter & Target & Shadow & ClutterTarget & TargetShadow & ShadowClutter & Training & Test &  \\ \midrule
A-ConvNet & 6.56\% (0.86) & \textbf{69.61\% (9.22)} & 23.82\% (3.16) & 0.72 & \textbf{1.75} & 0.35 & 97.24\% & 90.30\% & 0.30M \\
AM-CNN & 11.23\% (0.96) & \textbf{63.77\% (5.46)} & 25.01\% (2.14) & 1.57 & \textbf{3.40} & 0.69 & 100.0\% & 90.50\% & 2.60M \\
EfficientNet-B0 & 24.17\% (2.81) & \textbf{40.37\% (4.70)} & 35.47\% (4.12) & 1.13 & \textbf{1.58} & 0.36 & 99.12\% & 77.56\% & 4.02M \\
EfficientNet-B1 & 23.91\% (2.83) & \textbf{44.66\% (5.23)} & 31.43\% (3.71) & 1.36 & \textbf{1.83} & 0.67 & 99.21\% & 83.12\% & 6.53M \\
MVGGNet & 22.52\% (4.94) & \textbf{42.19\% (9.25)} & 35.29\% (7.74) & -0.64 & \textbf{3.96} & 1.46 & 100.0\% & 90.82\% & 16.81M \\
ResNet34 & 36.85\% (3.88) & \multicolumn{1}{l}{\textbf{37.44\% (3.94)}} & \multicolumn{1}{l}{25.71\% (2.71)} & \textbf{2.47} & 1.94 & 1.76 & 99.83\% & 86.51\% & \multicolumn{1}{l}{21.29M} \\
ResNet50 & 36.31\% (4.13) & \textbf{38.47\% (4.38)} & 25.21\% (2.87) & \textbf{1.81} & 1.79 & 1.76 & 99.97\% & 93.29\% & 23.53M \\
ConvNeXtTiny & \textbf{45.32\% (4.84)} & 30.44\% (3.25) & 24.24\% (2.59) & \textbf{1.08} & -0.44 & 1.26 & 99.88\% & 95.20\% & 27.83M \\ \bottomrule
\end{tabular}
}
\end{table*}